\pdfoutput=1
\documentclass[11pt]{article}

\usepackage[final]{coling}

\usepackage{times}
\usepackage{latexsym}

\usepackage[T1]{fontenc}
\usepackage[utf8]{inputenc}

\usepackage{microtype}

\usepackage{inconsolata}

\usepackage{graphicx}

\usepackage{amssymb}
\usepackage{amsmath}
\usepackage{amsthm}
\usepackage{booktabs}
\usepackage{enumitem}
\usepackage{color}
\usepackage{multirow}
\usepackage{booktabs}
\usepackage{geometry}
\usepackage{mdframed}
\usepackage{colortbl}
\usepackage{comment}
\usepackage{courier}
\usepackage{rotating}
\usepackage{tabularray}
\usepackage{soul}

\usepackage{makecell}
\usepackage{array}
\usepackage{graphicx}
\usepackage{ragged2e}

\usepackage{booktabs}
\usepackage{caption}
\usepackage{multirow}
\usepackage{adjustbox}

\newtheorem{definition}{Definition}

\newcommand{\suhang}[1]{\textcolor{blue}{SW: #1}}

\title{Exploring Language Model Generalization in Low-Resource Extractive QA}

\author{
  \textbf{Saptarshi Sengupta\textsuperscript{1}},
  \textbf{Wenpeng Yin\textsuperscript{1}},
  \textbf{Preslav Nakov\textsuperscript{2}},
  \textbf{Shreya Ghosh\textsuperscript{3}},
  \textbf{Suhang Wang\textsuperscript{1}} 
\\
  \textsuperscript{1}Pennsylvania State University, USA \\
 \textsuperscript{2}Mohamed bin Zayed University of Artificial Intelligence, UAE  \\
\textsuperscript{3}Indian Institute of Technology (IIT) Bhubaneswar, India  \\
   \{\texttt{sks6765, wenpeng, szw494}\}\texttt{@psu.edu} \\ \texttt{preslav.nakov@mbzuai.ac.ae, shreya@iitbbs.ac.in}
}

\begin{document}
\maketitle
\begin{abstract}
In this paper, we %question whether Large Language Models (LLMs) have bridged the generalization gap i.e. capabile of being deployed on any domain without performance drop. We 
investigate Extractive Question Answering (EQA) with Large Language Models (LLMs) under \textit{domain drift}, i.e., 
can LLMs generalize to domains that require specific knowledge such as medicine and law in a zero-shot fashion without additional in-domain training?~To this end, we devise a series of experiments to explain the performance gap empirically. Our findings suggest that: (a) LLMs struggle with dataset demands of closed domains such as retrieving long answer spans; (b) Certain LLMs, despite showing strong overall performance, display weaknesses in meeting basic requirements as discriminating between domain-specific senses of words which we link to pre-processing decisions; (c) Scaling model parameters is not always effective for cross-domain generalization; and (d) Closed-domain datasets are quantitatively much different than open-domain EQA datasets and current LLMs struggle to deal with them. Our findings point out important directions for improving existing LLMs. %to potentially bridge the generalization gap.
\end{abstract}

\section{Introduction}
\label{sec:Introduction}
%\suhang{the introduction not only introduces the problem you study, but also need to motivate why this is an important and novel problem, what's the novelty of our approach, and what our contributions are. However, the novelty and contribution are not clear for the current version - addressed}
%\suhang{It seems that many models tested in this paper are not considered as LLM, e.g., BERT and RoBERTa. They seem to be small language models. Shall we change the title and focus to LM-driven instead of LLM-driven - addressed}

For all their success in general-domain tasks, LLM performance in critical (or closed) reasoning domains such as medicine \cite{ullah2024challenges, informatics11030057} and law \cite{lai2023large} has been shown to be lacking, even on traditional tasks such as Natural Language Inference \cite{wang-etal-2024-rethinking}. \ul{This is the first focus of our paper, i.e., examining the reasons for the poor performance of language models in closed domains}.

Our examination focuses on Extractive Question Answering (EQA) (\S \ref{sec:Problem_Formulation}), i.e., the task of retrieving a contiguous span of tokens from a passage of text to answer a query based on it. In closed domains, response quality is crucial. Unfortunately, as generative models are prone to hallucination \cite{huang2023survey} or sensitive to the location of the answer span \cite{liu2024lost}, they cannot be reliably used (yet) in such domains \cite{magesh2024hallucination, pal-etal-2023-med}. As such, extractive retrieval offers better trust in the model response. This is because a model does not need to create new information, but rather locate gold annotated text spans. \ul{This is the second focus of this paper, i.e., studying EQA in closed domains.}

Self-supervised pre-training on in-domain data is generally utilized as the strategy for garnering domain expertise. However, for esoteric subjects, large-scale training data is seldom available. For example, the corpus curated by \citet{bhattacharjee2024indus} discussing among others, \textit{astrophysics} literature, consists of only 66B tokens, a small fraction of the 2T token corpus used by Llama 2 \cite{touvron2023llama2}, a general domain model. As such, it is not always possible to perform in-domain pre-training. However, after pre-training in the general domain, a model can be trained to work well for related EQA (c.f. Fig. \ref{fig:problem_statement}). This leads us to the \ul{final focus of this paper, i.e., without additional in-domain fine-tuning, we investigate the extent to which language models can generalize (zero-shot) for closed-domain EQA.}

Overall, our main contributions are, (i) We motivate the importance of EQA and the challenges associated with cross-domain generalization by highlighting the poor performance of current models, (ii) Through various experiments, we offer insights into the limitations of current EQA models and complexities of closed-domain datasets that need to be addressed for adaptation across domains, (iii) Finally, we provide recommendations on model usage for EQA in particular, which can be leveraged for other tasks as well.

\section{Problem Formulation}
\label{sec:Problem_Formulation}
%\suhang{generally, people don't put the formal problem definition in the introduction. the introduction should focus more on the motivation of the problem, why it is a novel, challenging and important problem, and what's our contribution. In the introduction, you can briefly explain the EQA task and explain why we choose this task to investigate the generalization ability of LLMs. You can put the formal EQA task in the next section - addressed}

EQA has three components \cite{liu2019neural} I) Context ($C$): The passage on which the question is based and from which the answer must be drawn; II) Question ($Q$): The query based on the context; III) Answer ($A$): The span of context tokens which answers the question. Formally, EQA is defined as,
\begin{definition}
    Given $n$ tokens, $t$, as context $C = \{t_1, \ldots t_n \}$ and question $Q$, EQA aims to extract a continuous subsequence of $k$ tokens from the context as the answer $A$, %\suhang{should be xxx, i.e., xxx. do not omit the comma} 
    i.e., $A = \{t_i, \ldots t_{i+k} \}$ where $1 \leq i \leq (i+k) \leq n$. In other words, the aim is to learn the function $EQA \colon f(C, Q) \rightarrow A$
\end{definition}

\begin{figure}
    \centering
    \includegraphics[scale=0.7]{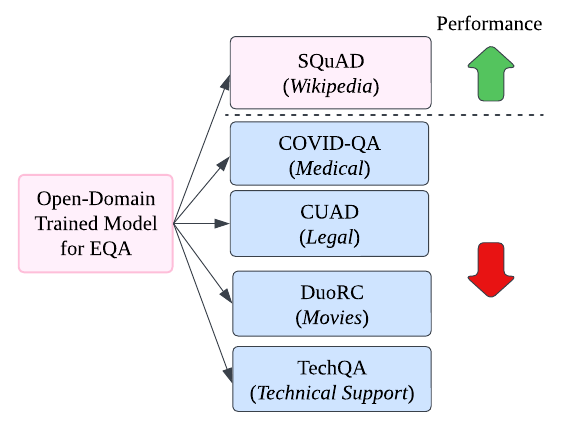}
    \vskip -1em
    \caption{We attempt to explain the performance drop when a model is trained using in-domain (ID) datasets (SQuAD; pink) and tested on ID data (SQuAD) v/s OOD (out-of-domain) data (blue).}
    \label{fig:problem_statement}
\end{figure}

Figure \ref{fig:problem_statement} explains our problem statement. Usually, models are trained on EQA datasets that align with their pre-training data, generally web or open-domain corpora such as Wikipedia. This leads to strong performance when the test set is from the same domain. However, if the domain of the test set is misaligned with the training data, performance degrades sharply. We aim to study why this decline takes place. We test models that are trained on an ID EQA dataset viz., SQuAD \cite{rajpurkar2016squad} and test them on four OOD datasets, DuoRC \cite{saha-etal-2018-duorc}, CUAD \cite{hendrycks2021cuad}, COVID-QA \cite{moller-etal-2020-covid} and TechQA \cite{castelli-etal-2020-techqa} without further training (zero-shot) and explain the performance gap.

\subsection{Evaluating EQA models}

Metrics used for evaluating an EQA model are EM (Exact Match) and F1. EM looks for a verbatim match between the predicted and gold answer and, is thus a 0/1 measure. F1 calculates the harmonic mean of the prediction precision (count of shared words between the prediction and gold span/count of words in the prediction) and recall (count of common words/count of words in the gold span).

EM and F1 have a very low tolerance for error due to relying on token overlap and are thus, strict measures. Despite that, prior work has primarily utilized them for reporting scores. This is because creating better, and more nuanced metrics is a non-trivial task. While there have been attempts to this end such as BERTScore \cite{DBLP:conf/iclr/ZhangKWWA20} and TigerScore \cite{jiang2024tigerscore}, these have not yet been widely adopted for EQA. Furthermore, there are active studies \cite{farea2024experimental} investigating the impact of EM/F1, which shows the preference of studies to favour a simpler metric over complex measures.

Assuming we have a better measure, isolating the correct portion of the model's generation (as the answer span) is another challenge. As shown in Figure \ref{fig:gemma}, LLMs produce answers in a variety of formats. Automatically identifying where the correct (answer) sentence lies is again a non-trivial task. As a consequence of this, many works \cite{labrak2024biomistral, han2023medalpaca, chen2023meditron} focus on multiple-choice QA as the generated text is easier to parse.

%Figure  gives an overview of Transfer Learning and what aspect of it we study in this paper. %A randomly initialized model undergoes self-supervised pre-training on a corpus to learn a general overview of language and the respective domain. Next, it is fine-tuned for a certain task. Following this, it can either be further trained or deployed as is on the target dataset. %through various instruction following means \cite{zhang2023instruction, lou2024comprehensive}. 
%We are interested in \textit{domain adaptation}, i.e., keeping the task same, how well does a model perform on a domain much different from its training domain; and \textit{zero-shot generalization}, i.e., not involving further training. 

%\suhang{we should clearly write down our novelty/significant findings/contributions. I cannot get these from the introduction - addressed}

%%%%%%%%%%%%%%%%%%%%%%%%%%%%%%%%%%%%%%%%%%%%%%%%%%%%%%%%%%%%%%%%%%%%%%%%

\section{Experiments and Results}

\begin{figure}
    \centering
    \includegraphics[scale=0.45]{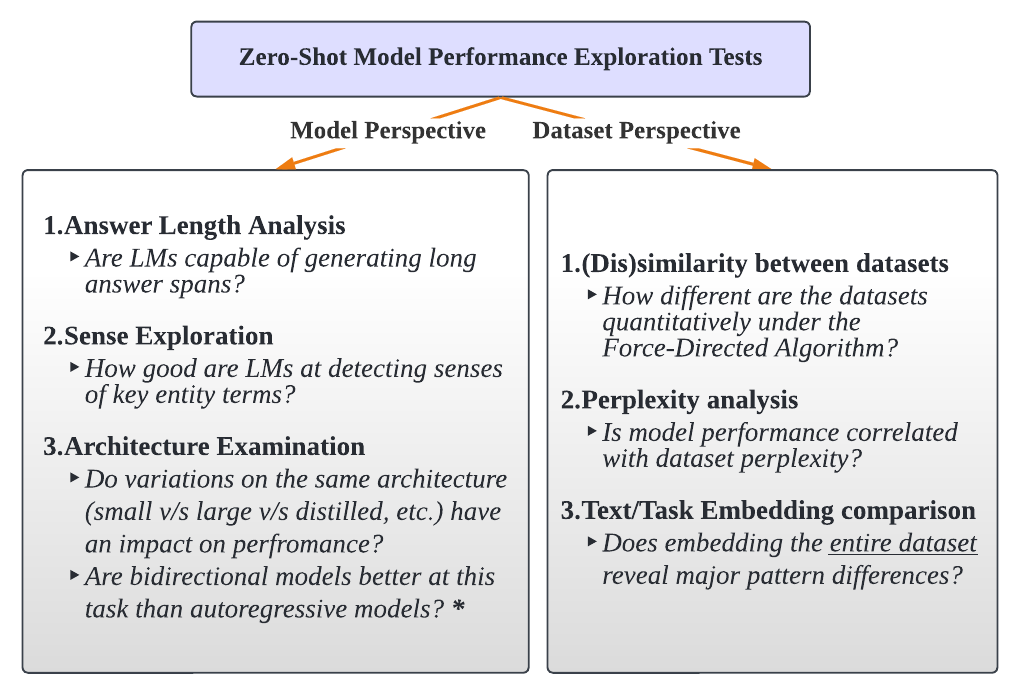}
    \caption{Proposed Experiments. *We provided a detailed analysis of causal LLMs in Appendix \ref{app:Autoregressive Models} and discuss why they are suboptimal for EQA.}
    \vskip -1em
    \label{fig:exp}
\end{figure}

We %experiments that are performed for exploring the generalization capabilities of LMs for zero-shot EQA. We  
classify our experiments (Figure \ref{fig:exp}) as, \textbf{Model Perspective}, i.e., looking at limitations in the model themselves and \textbf{Dataset Perspective}, i.e., examining the complexities of the OOD datasets.

    %\item \textbf{Model Perspective}: Here, we examine limitations in the model architecture for the performance drop such as being unable to generate \textit{long answer spans}, discriminate between multiple senses (\textit{polysemy}) of entities etc.
    
    %\item \textbf{Dataset Perspective}: %Here we look at characteristics of the datasets themselves and their impact on performance. 
    %We recognize that the performance drop might also be attributed to the specialized language of closed-domains. With experiments such as correlating model performance with \textit{perplexity} we identify issues in OOD datasets that restrict performance.
    
    %have very  resulting in linguistic divergence from their ID counterparts. 

Each experiment, under model-perspective\footnote{We unify the discussion of the dataset-perspective experiments as they collectively describe a common story.} is structured to answer the I) \textit{Hypothesis} (in \textcolor{blue}{blue}) - What is the main idea being investigated? II) \textit{Motivation} - What is the background/reason for performing this test? III) \textit{Experiment Setting} - How do we test the hypothesis? IV) \textit{Findings} - What are the results of experiments? V) \textit{Key Takeaways} - What are the main lessons learned from the experiment?

\subsection{Models and Datasets}

We test various architectures for EQA, categorized as (i) Non-transformer based, BiDAF \cite{seo2017bidirectional} and QANet \cite{DBLP:conf/iclr/YuDLZ00L18}; (i) Transformer-based further categorized as encoder-based, including BERT (and its variants), RoBERTa \cite{liu2019roberta}, and decoder-based, including Falcon \cite{almazrouei2023falcon}, Platypus \cite{lee2023platypus}, Gemma \cite{team2024gemma}, Mistral \cite{jiang2023mistral} \textit{inter alia}. 

Overall, we use five datasets in this study covering general knowledge (SQuAD), COVID-related medical literature (COVID-QA), legal documentation (CUAD), pop-culture and movies (DuoRC) and technical customer support (TechQA). Additional details on the models and datasets are provided in Appendix \ref{app:Dataset_Models}.

\subsection{Model Perspective}

Under this category of experiments, we look at different aspects of a model to determine potential architectural limitations that lead to their poor performance in closed-domain EQA. 

\subsubsection{Predicted Answer Length Analysis}
\label{sec:answer_length}
%\suhang{I do not think we want to put the hypothesis under the subsubsection title. We can put in motivation and highlight it there - addressed}
\paragraph*{Motivation} We hypothesize that \textcolor{blue}{\textit{current EQA models are weak in generating long answer spans matching the distribution of the gold data answer spans}}. Closed-domain datasets have longer questions, contexts and answers than SQuAD (c.f. \ref{tab:dataset-stats}). Thus, to answer their questions, a model needs to produce longer spans of text typically not required for simple factoid-based questions found in SQuAD. %This is a reasonable ask considering we would expect our systems to produce detailed explanations for involved questions such as \textit{What is the main cause of HIV-1 infection in children?} (COVID-QA) as opposed to simpler questions as \textit{What team was the winner of Super Bowl XXXIII?} (SQuAD). 
This leads to our hypothesis that \textit{model performance is impacted due to the inability to produce long answer spans}. In other words, we test whether EQA models overfit the average gold answer length in the training data.

\paragraph*{Experiment Setting} We examine if architectures specifically trained for EQA still suffer from the same-length generalization drawback. We test if the issue persists for both non-transformer (BiDAF, QANet) and Transformer-based Masked Language Models (MLM) (BERT, RoBERTa). %This enables us to understand whether this is a problem with Transformers or with neural networks in general. 
Causal LMs (CLM) are not used here as during inference, we can %\suhang{cannot?} 
control generation till the window limit, giving them the flexibility to produce shorter or longer spans.

%\suhang{I think this measure might have some issues. For example, assume we only have two test samples, one has ground truth length as 20, the other one has 30. If the model gives lengths as 30 and 20, respectively, based on your calculation, $\Delta = 25 - 25 =0$. Is that what we want? Is there some better measure? - addressed - yes sir, that is fine. As long as we have a non-negative difference i.e. >= 0, it indicates that the model is either matching or exceeding the gold span average. We do look at the overall distribution of prediction lengths w.r.t gold in the appendix.}

To test our hypothesis, we determine the \textit{average number of characters} in the predicted answer spans and calculate the difference between it and the average gold span for the given datasets\footnote{We also look at the distribution of lengths in Appendix \ref{app:Categorical_Answer_Length_Analysis}.}. We use characters instead of tokens to have a consistent scheme across models, as each model uses their own tokenization. A non-negative difference indicates that a model produces spans matching or exceeding the expected gold length and vice versa.

\begin{table}[t]
  \centering
  \resizebox{\columnwidth}{!}{\begin{tabular}{llccccc}
    \toprule
    \multirow{2}{*}{Domain} & \multirow{2}{*}{Model} & \multicolumn{3}{c}{Avg. Val. \#chars} & \multirow{2}{*}{EM} & \multirow{2}{*}{F1} \\
    \cmidrule(r){3-5}
    & & True & Predicted & $\Delta$ & \\
    \midrule   
    \multirow{4}{*}{SQuAD (Open/General)} & BiDAF & \multirow{4}{*}{18.73} & 25.31 & 6.58 & 65.73 & 75.98 \\
    & QANet & & 23.74 & 5.01 & 26.3 & 36.81  \\
    & BERT & & 18.18 & -0.55 & 80.95 & 88.25 \\
    & RoBERTa & & 18.03 & -0.7 & 82.73	& 90.04  \\
    \midrule
    \multirow{4}{*}{COVID-QA (BioMedical)} & BiDAF & \multirow{4}{*}{93.42} & 986.73 & 893.31 & 17.43 & 38.3 \\
    & QANet & & 460.81 & 367.39 & 0.99 & 5.76 \\
    & BERT & & 28.81 & -64.61 & 22.39 & 42.11 \\
    & RoBERTa & & 25.81 & -67.61 & 21.89 & 40.2 \\
    \midrule
    \multirow{4}{*}{CUAD (Law)} & BiDAF & \multirow{4}{*}{120.07} & 5261.19 &  5141.12 & 5.06 & 16.81 \\
    & QANet & & 277.32 & 157.25 & 0.8  & 7.01 \\
    & BERT & & 33.55 & -86.52 & 7.72 & 15 \\
    & RoBERTa & & 19.55 & -100.52 & 4.02 & 7.7 \\
    \midrule
    \multirow{4}{*}{DuoRC (Movie Plots)} & BiDAF & \multirow{4}{*}{14.27} & 66.23 & 51.96 & 43.99 & 56.53 \\
    & QANet & & 155.32 & 141.05 & 19.03 & 27.56 \\
    & BERT & & 14.72 & 0.45 & 55.59 & 69.25 \\
    & RoBERTa & & 14.03 & -0.24 & 60.6 & 74.43 \\
    \midrule
    \multirow{4}{*}{TechQA (Technical QA)} & BiDAF & \multirow{4}{*}{156.79} & 4302.93 & 4146.14 & 0.625 & 14.56 \\
    & QANet & & 387.2 & 230.41 & 0 & 7.65 \\
    & BERT & & 18.42 & -138.37 & 0 & 9.19 \\
    & RoBERTa & & 26.89 & -129.9 & 0.625 & 5.94 \\
    \bottomrule
  \end{tabular}}
  \caption{Zero-Shot Performance in Different Domains. $\Delta =$ Average (Predicted - Gold) Answer Span Length}
  \label{tab:ans-length}
  \vskip -1em
\end{table}

%\suhang{in the table, you also report EM and F1. 
%1. How do we calculate EM and F1 for the EQA task? I think we need to explain it at begining. EQA doesn't seem to be a mainstream task. Not many people will know it - addressed now. 
%2. Since you put the results in terms of EM and F1, can you also discuss some observations based on these two measures, maybe together with $\Delta$ - done (321)}

\paragraph*{Findings} From Table \ref{tab:ans-length}, we see how well BERT and RoBERTa approximate the average gold answer length of SQuAD. %This, coupled with the fact that they are superior architectures, leads to the best performance on this dataset. However, the same cannot be said for the two on the other datasets. 
However, on the OOD datasets, both consistently produce shorter spans leading to negative $\Delta$. %between the predicted and gold span lengths on each dataset 
%(BERT on DuoRC being the only exception). 
BERT breaks the 30-length mark only once (for CUAD) while RoBERTa can barely go beyond 25 characters. Interestingly, we see that BERT produces longer spans for all of the datasets except TechQA. We know that RoBERTa is trained on a much larger corpus and for a longer number of epochs than BERT. Producing consistently smaller spans indicates overfitting on the training corpus. Building on this analysis, we see that apart from SQuAD and DuoRC, BERT performs much better than RoBERTa on COVID-QA, CUAD and TechQA. %While fascinating, this result makes sense seeing as SQuAD and DuoRC are both sourced from Wikipedia which is again found in RoBERTa's pre-training corpus. 
Although BERT is also trained on Wikipedia, it is considered \textit{undertrained} w.r.t RoBERTa and hence shows inferior performance on those two datasets but better performance in more complex domains.

For the non-transformer models, the clear winner among the two is BiDAF which % as QANet does not even come close to its performance. Additionally, 
produces much longer spans than QANet. BiDAF, %has 2.6M parameters whereas RoBERTa\textsubscript{BASE} has 125M parameters. 
despite being much smaller than RoBERTa, outperforms it on CUAD and TechQA and comes close to it on COVID-QA (in terms of F1). This shows that \textit{larger models do not necessarily equate to better cross-domain generalization}.

BiDAF generates much longer answer spans compared to other models. These longer spans cover more context, which helps improve the F1 score as it accounts for overlapping tokens. In domains requiring detailed answers, such as law (CUAD) and technical customer support (TechQA), longer spans are beneficial because complex questions benefit from more thorough responses.

\paragraph*{Key Takeaways} Here, we observe: (i) Smaller models like BiDAF display competitive performance in zero-shot EQA over larger and more capable models. Thus, \textit{it is not guaranteed that scale equates with domain requirements such as generating longer predictions}; and (ii) When using Transformers, \textit{we should start with BERT as it displays a better tendency to learn new domains}.

% \begin{enumerate}[leftmargin=*]
%     \item Smaller models like BiDAF display competitive performance in zero-shot EQA over larger and more capable models. Thus, \textbf{it is not guaranteed that scale equates with domain requirements such as generating longer predictions}.

%     \item When using Transformers, \textbf{we should start with BERT as it displays a better tendency to learn new domains}.
% \end{enumerate}

\subsubsection{Examining Polysemy of Domain Terms}
\label{sec:Examining Polysemy of Domain Terms}

\paragraph*{Motivation} We hypothesize that \textcolor{blue}{\textit{LLMs are weak in detecting senses of relevant domain terms}}. Polysemy is a linguistic phenomenon to describe words that take on multiple meanings or \textit{senses}, e.g., %\suhang{e.g. means for example, you either use for example, or e.g.,} 
\textit{bank} can mean either a financial institution or the portion of land beside a river. For closed-domains datasets, we reason that words with multiple meanings will, on average, show only their \textit{expected} usage for the domain. E.g. we would expect that a term such as \textit{Party} would only take on the \textit{group} meaning rather than the \textit{occasion} meaning for a legal QA dataset (CUAD). Additionally, considering the small number of samples in closed-domain datasets, as opposed to their open-domain counterparts, it would be difficult to expect a dataset on technical customer support (TechQA) to have many/any instances of the \textit{coffee} sense for \textit{Java}. Taking this into account, \textit{we test whether LLMs can discriminate between ID and OOD senses of polysemous domain terms}.

%Non-contextualized models such as Word2Vec \cite{mikolov2013distributed} and GloVe \cite{pennington-etal-2014-glove} failed to capture polysemy as they condensed all senses of a word into a single embedding. With the advent of contextualized language models, word embeddings were computed on the fly based on the surrounding context thus providing a more sophisticated approach to handle polysemy. 

%(although this may not always be the case as in \textit{cold} for \textit{temperature} and \textit{condition} for biomedical datasets). 

\paragraph*{Experiment Setting}

To test this hypothesis, we create a small dataset of polysemous domain terms that appear in the vocabulary of various contextualized (MLM/CLM) models and the respective datasets along with their associated contexts. We start by tokenizing the contexts of the training split for each dataset and filter out tokens absent from the model's vocabulary, stop words, numbers and punctuation. Sorting the filtered list by frequency, we randomly select five polysemous terms relevant to the domain and retrieve their contexts.

As expected, the datasets usually show only a single sense of a word, and as such we had to rely on external resources to obtain contexts for the other senses. To this end, we manually scrape a well-known website for reliable word definitions and usages, i.e. \texttt{vocabulary.com}. In total, we had ten contexts per sense of a given word. 

%We lowercase each word and context to be able to use the word in its untokenized form as we want to examine whole word and not subword embeddings. For e.g. \textit{Agreement} appears as [\textunderscore{}\textit{Ag}, \textit{re}, \textit{ement}] in the vocabulary for Platypus but \textit{\textunderscore{}}\textit{agreement} when lowercased. Next, 
We run each context through the frozen models and extract contextualized embeddings for the polysemous words in our dataset. Using these embeddings, we compute the average cosine similarity between the target word from the same and different sense groups. The overall \textit{logic here is that intra/same-group similarity is expected to be higher, while inter/different-group would be lower}. If not, this can indicate that the models are incapable, to an extent, of discriminating between domain and non-domain senses of words, which in turn contributes to their poor generalization.

\begin{table*}
\centering
\Large
% COVID-QA
\resizebox{\linewidth}{!}{%
\begin{tabular}{|c|c|cccccc|ccc|ccc|ccc|ccc|} \Xhline{1mm}
Dataset & Word & \multicolumn{6}{c|}{Cell} & \multicolumn{3}{c|}{Expression} & \multicolumn{3}{c|}{Study} & \multicolumn{3}{c|}{Treatment} & \multicolumn{3}{c|}{Host} \\ \hline
\multirow{13}{*}{\rotcell{\makebox[0pt][l]{COVID-QA}}} & Model & \textit{bio (s1)} & \textit{room (s2)} & \textit{phone (s3)} & \textit{(s1, s2)} & \textit{(s1, s3)} & \textit{(s2, s3)} & \textit{bio (s1)} & \textit{non-verbal (s2)} & \textit{(s1, s2)} & \textit{experiment (s1)} & \textit{room (s2)} & \textit{(s1, s2)} & \textit{medical care (s1)} & \textit{behavior (s2)} & \textit{(s1, s2)} & \textit{organism (s1)} & \textit{organizer (s2)} & \textit{(s1, s2)} \\ \cline{2-20}
 & BERT & 0.75 & 0.73 & 0.6 & 0.37 & 0.4 & 0.38 & 0.83 & 0.65 & 0.37 & 0.8 & 0.66 & 0.41 & 0.73 & 0.59 & 0.39 & 0.81 & 0.61 & 0.42 \\
 & RoBERTa & 0.96 & 0.95 & 0.91 & 0.89 & 0.87 & 0.89 & 0.97 & 0.93 & 0.87 & 0.96 & 0.96 & 0.89 & 0.95 & 0.92 & 0.87 & 0.96 & 0.94 & 0.88 \\
 & BioBERT & 0.81 & 0.83 & 0.85 & 0.71 & 0.74 & 0.77 & 0.88 & 0.82 & 0.74 & 0.85 & 0.88 & 0.76 & 0.84 & 0.92 & 0.78 & 0.87 & 0.84 & 0.76 \\
 & SciBERT & 0.73 & 0.77 & 0.76 & 0.57 & 0.62 & 0.66 & 0.8 & 0.76 & 0.59 & 0.77 & 0.82 & 0.65 & 0.76 & 0.85 & 0.64 & 0.77 & 0.81 & 0.61 \\
 & SenseBERT & 0.88 & 0.89 & 0.82 & 0.75 & 0.73 & 0.73 & 0.92 & 0.86 & 0.75 & 0.9 & 0.89 & 0.79 & 0.87 & 0.88 & 0.76 & 0.92 & 0.86 & 0.76 \\
 & Falcon & 0.98 & 0.98 & 0.99 & 0.98 & 0.98 & 0.98 & 0.98 & 0.98 & 0.98 & 0.98 & 0.98 & 0.97 & 0.98 & 0.98 & 0.97 & 0.99 & 0.97 & 0.97 \\
 & Platypus & 0.55 & 0.53 & 0.67 & 0.32 & 0.31 & 0.46 & 0.56 & 0.5 & 0.33 & 0.63 & 0.48 & 0.4 & 0.6 & 0.54 & 0.42 & 0.58 & 0.41 & 0.23 \\
 & Gemma & 0.7 & 0.86 & 0.86 & 0.66 & 0.71 & 0.83 & 0.87 & 0.86 & 0.83 & 0.92 & 0.77 & 0.79 & 0.85 & 0.93 & 0.83 & 0.82 & 0.88 & 0.77 \\
 & Mistral & 0.43 & 0.53 & 0.52 & 0.2 & 0.25 & 0.42 & 0.54 & 0.53 & 0.22 & 0.53 & 0.4 & 0.25 & 0.45 & 0.51 & 0.27 & 0.5 & 0.45 & 0.14 \\ \hline
\end{tabular}
}

\hspace{1cm}
%CUAD
\resizebox{\linewidth}{!}{%
\begin{tabular}{|c|c|ccc|ccc|ccc|ccc|ccc|} 
\Xhline{1mm}
Dataset & Word & \multicolumn{3}{c|}{Party} & \multicolumn{3}{c|}{Agreement} & \multicolumn{3}{c|}{Company} & \multicolumn{3}{c|}{Product} & \multicolumn{3}{c|}{Notice} \\ \hline
\multirow{12}{*}{\rotcell{\makebox[0pt][l]{CUAD}}} & Model & \textit{group (s1)} & \textit{occasion (s2)} & \textit{(s1, s2)} & \textit{contract (s1)} & \textit{understanding (s2)} & \textit{(s1, s2)} & \textit{organization (s1)} & \textit{companionship (s2)} & \textit{(s1, s2)} & \textit{goods (s1)} & \textit{consequence (s2)} & \textit{(s1, s2)} & \textit{announcement (s1)} & \textit{observation (s2)} & \textit{(s1, s2)} \\ \cline{2-17}
 & BERT & 0.69 & 0.67 & 0.34 & 0.83 & 0.56 & 0.52 & 0.73 & 0.58 & 0.36 & 0.68 & 0.67 & 0.41 & 0.72 & 0.67 & 0.38 \\
 & RoBERTa & 0.9 & 0.95 & 0.84 & 0.98 & 0.91 & 0.84 & 0.95 & 0.91 & 0.82 & 0.91 & 0.94 & 0.82 & 0.91 & 0.93 & 0.84 \\
 & FinBERT & 0.75 & 0.65 & 0.37 & 0.84 & 0.6 & 0.52 & 0.79 & 0.47 & 0.39 & 0.59 & 0.71 & 0.42 & 0.75 & 0.7 & 0.4 \\
 & LegalBERT & 0.66 & 0.78 & 0.56 & 0.83 & 0.71 & 0.62 & 0.84 & 0.73 & 0.69 & 0.78 & 0.79 & 0.64 & 0.78 & 0.8 & 0.64 \\
 & SenseBERT & 0.86 & 0.8 & 0.7 & 0.95 & 0.82 & 0.83 & 0.9 & 0.73 & 0.71 & 0.8 & 0.9 & 0.75 & 0.89 & 0.9 & 0.77 \\
 & Falcon & 0.98 & 0.98 & 0.97 & 0.98 & 0.98 & 0.98 & 0.97 & 0.98 & 0.97 & 0.98 & 0.97 & 0.97 & 0.97 & 0.99 & 0.97 \\
 & Platypus & 0.57 & 0.62 & 0.4 & 0.67 & 0.63 & 0.56 & 0.54 & 0.58 & 0.17 & 0.47 & 0.57 & 0.36 & 0.7 & 0.61 & 0.38 \\
 & Gemma & 0.83 & 0.94 & 0.84 & 0.88 & 0.96 & 0.9 & 0.84 & 0.96 & 0.85 & 0.64 & 0.85 & 0.72 & 0.92 & 0.9 & 0.89 \\
 & Mistral & 0.41 & 0.55 & 0.17 & 0.52 & 0.55 & 0.35 & 0.41 & 0.56 & 0.26 & 0.34 & 0.53 & 0.24 & 0.56 & 0.62 & 0.2 \\ \hline
\end{tabular}
}

\hspace{1cm}
%TechQA
\resizebox{\linewidth}{!}{%
\begin{tabular}{|p{0.3cm}|c|ccc|cccccc|ccc|ccc|ccc|} 
\Xhline{1mm}
\multicolumn{1}{|c|}{Dataset} & Word & \multicolumn{3}{c|}{Server} & \multicolumn{6}{c|}{Application} & \multicolumn{3}{c|}{Following} & \multicolumn{3}{c|}{Java} & \multicolumn{3}{c|}{Windows} \\ \hline
\multirow{9}{*}{\rotcell{TechQA}} & Model & \textit{host (s1)} & \textit{waiter (s2)} & \textit{(s1, s2)} & \textit{program (s1)} & \textit{request (s2)} & \textit{use (s3)} & \textit{(s1, s2)} & \textit{(s1, s3)} & \textit{(s2, s3)} & \textit{reference (s1)} & \textit{pursue (s2)} & \textit{(s1, s2)} & \textit{software (s1)} & \textit{coffee (s2)} & \textit{(s1, s2)} & \textit{software (s1)} & \textit{framework (s2)} & \textit{(s1, s2)} \\ \cline{2-20}
 & BERT & 0.78 & 0.8 & 0.48 & 0.71 & 0.72 & 0.53 & 0.4 & 0.35 & 0.41 & 0.61 & 0.58 & 0.31 & 0.84 & 0.67 & 0.48 & 0.77 & 0.71 & 0.33 \\
 & RoBERTa & 0.94 & 0.95 & 0.89 & 0.93 & 0.95 & 0.92 & 0.87 & 0.86 & 0.89 & 0.91 & 0.92 & 0.84 & 0.94 & 0.91 & 0.84 & 0.93 & 0.94 & 0.79 \\
 & SciBERT & 0.79 & 0.82 & 0.71 & 0.75 & 0.76 & 0.74 & 0.62 & 0.59 & 0.66 & 0.67 & 0.69 & 0.55 & 0.79 & 0.65 & 0.59 & 0.71 & 0.85 & 0.62 \\
 & SenseBERT & 0.88 & 0.89 & 0.75 & 0.86 & 0.85 & 0.81 & 0.74 & 0.71 & 0.73 & 0.72 & 0.81 & 0.52 & 0.9 & 0.69 & 0.53 & 0.83 & 0.92 & 0.76 \\
 & Falcon & 0.98 & 0.98 & 0.97 & 0.97 & 0.98 & 0.98 & 0.97 & 0.97 & 0.98 & 0.99 & 0.98 & 0.97 & 0.99 & 0.98 & 0.98 & 0.99 & 0.98 & 0.98 \\
 & Platypus & 0.63 & 0.4 & 0.32 & 0.52 & 0.55 & 0.46 & 0.16 & 0.16 & 0.45 & 0.6 & 0.45 & 0.22 & 0.77 & 0.35 & 0.17 & 0.73 & 0.42 & 0.16 \\
 & Gemma & 0.79 & 0.71 & 0.72 & 0.61 & 0.85 & 0.91 & 0.61 & 0.63 & 0.86 & 0.55 & 0.83 & 0.59 & 0.8 & 0.83 & 0.79 & 0.93 & 0.78 & 0.8 \\
 & Mistral & 0.5 & 0.41 & 0.26 & 0.44 & 0.5 & 0.43 & 0.25 & 0.19 & 0.35 & 0.66 & 0.58 & 0.32 & 0.59 & 0.34 & 0.28 & 0.57 & 0.45 & 0.23 \\ \hline
\end{tabular}
}

\hspace{1cm}
%DuoRC

\resizebox{\linewidth}{!}{%
\begin{tabular}{|p{1.3cm}|c|ccc|ccc|ccc|ccc|ccc|} \Xhline{1mm}
Dataset & Word & \multicolumn{3}{c|}{Film} & \multicolumn{3}{c|}{Shoot} & \multicolumn{3}{c|}{Hand} & \multicolumn{3}{c|}{Couple} & \multicolumn{3}{c|}{Past} \\ \hline
\multirow{8}{*}{\rotcell{DuoRC}} & Model & \textit{movie (s1)} & \textit{verb (s2)} & \textit{(s1, s2)} & \textit{bullet (s1)} & \textit{record (s2)} & \textit{(s1, s2)} & \textit{anatomy (s1)} & \textit{situation (s2)} & \textit{(s1, s2)} & \textit{pair (s1)} & \textit{few (s2)} & \textit{(s1, s2)} & \textit{time (s1)} & \textit{pass (s2)} & \textit{(s1, s2)} \\ \cline{2-17}
 & BERT & 0.73 & 0.65 & 0.5 & 0.64 & 0.59 & 0.5 & 0.52 & 0.5 & 0.41 & 0.57 & 0.56 & 0.32 & 0.6 & 0.48 & 0.36 \\
 & RoBERTa & 0.97 & 0.93 & 0.9 & 0.95 & 0.93 & 0.92 & 0.92 & 0.85 & 0.85 & 0.94 & 0.94 & 0.87 & 0.93 & 0.89 & 0.86 \\
 & SenseBERT & 0.91 & 0.84 & 0.76 & 0.81 & 0.72 & 0.68 & 0.7 & 0.63 & 0.59 & 0.84 & 0.87 & 0.73 & 0.73 & 0.65 & 0.51 \\
 & Falcon & 0.98 & 0.99 & 0.97 & 0.98 & 0.97 & 0.97 & 0.98 & 0.98 & 0.98 & 0.98 & 0.97 & 0.97 & 0.99 & 0.99 & 0.98 \\
 & Platypus & 0.59 & 0.53 & 0.36 & 0.45 & 0.33 & 0.36 & 0.61 & 0.67 & 0.6 & 0.69 & 0.53 & 0.41 & 0.5 & 0.37 & 0.32 \\
 & Gemma & 0.87 & 0.68 & 0.73 & 0.79 & 0.76 & 0.75 & 0.96 & 0.97 & 0.96 & 0.79 & 0.85 & 0.75 & 0.97 & 0.76 & 0.73 \\
 & Mistral & 0.69 & 0.55 & 0.4 & 0.48 & 0.41 & 0.38 & 0.4 & 0.53 & 0.34 & 0.64 & 0.55 & 0.4 & 0.45 & 0.45 & 0.28 \\ \hline
\end{tabular}
}

\caption{Semantic Similarity Results. s(\textit{i}) indicates sense number and intra-group sense similarity. s(\textit{i,j}) indicates inter-group sense similarity. $\uparrow$ s(\textit{i}) + $\downarrow$ s(\textit{i,j}) indicates a stronger model.}
\label{tab:SS_Res}
\end{table*}

\paragraph*{Findings}

Firstly, from Table \ref{tab:SS_Res} we see an interesting connection between RoBERTa and Falcon. While both are different styles of models, each reports consistently high scores across all words and senses. Intra/inter-sense similarity scores for Falcon never falls below 0.97 for any dataset. This is concerning as it indicates that it fails to recognize differences in word usage across domains. While RoBERTa also shows higher similarity scores, it discriminates senses to a better extent (lower inter-sense scores).

%Noting that RoBERTa and Falcon are completely different models, 
We question if there is a common link between the two to explain the high similarity scores and find that both models rely on the same %\footnote{Although it is not mentioned in their paper, we see that Falcon performs similar tokenization as RoBERTa and hence make this claim.} 
tokenization scheme, i.e., byte-level BPE (Byte-Pair Encoding) an algorithm which treats individual bytes as tokens \cite{sennrich-etal-2016-neural}. On the other hand, Platypus, Mistral, and Gemma use better SentencePiece BPE \cite{kudo-richardson-2018-sentencepiece} which does not assume that words %\footnote{\url{https://huggingface.co/docs/transformers/en/tokenizer\_summary\#sentencepiece}} 
are space-separated.%, leading to better representations. 

The impact of tokenization on performance is a non-trivial issue particularly when dealing with out-of-vocabulary words as shown in \citet{soler2024impact} which extends to OOD senses. In fact, as shown by \citet{bostrom2020byte}, straight BPE %(character/byte-level) 
schemes are inferior to Unigram tokenization \cite{kudo-2018-subword} which in turn is not used in isolation, but coupled %\footnote{\url{https://huggingface.co/docs/transformers/en/tokenizer\_summary\#unigram}} 
with SentencePiece to form the basis for tokenizers for newer models. Additionally, it is shown by \citet{bostrom2020byte} that the latter mode of tokenization leads to better performance for QA tasks. All of this explains why Platypus, Gemma and Mistral show greater semantic awareness than Falcon and RoBERTa and by extension why Platypus, on average, outperforms Falcon across all datasets (c.f. Table \ref{tab:dec_zs}). %Although Gemma shares the same tokenization scheme as Platypus, it consistently shows higher inter/intra-sense similarity scores over Platypus. We link this reason to its poor performance across all of the datasets (c.f. Table \ref{tab:dec_zs}).
Finally, while encoders show higher similarity scores than decoders it does not imply that they are worse at sense discrimination. BERT consistently shows high intra-sense and low inter-sense similarity indicating a greater degree of sense separation. %in comparison to Gemma which occasionally shows higher inter-sense similarity than intra-sense.

\paragraph*{Key Takeaways}
The main findings here are: (i) Tokenization matters - Models employing \textit{SentencePiece + Unigram tokenization demonstrate overall better performance and sense discrimination}; %\suhang{consider using textit instead of textbf to highlight. The paragraph caption is already boldface. We have too many boldface words, making it difficult to read} 
(ii)  Despite their overall poor performance (c.f. Appendix \ref{app:Autoregressive Models}), \textit{CLMs exhibit a strong degree of sense awareness} in closed domains; (iii) \textit{BERT should be used as a starting point for linguistically challenging datasets}, as it shows strong sense discrimination in closed domains occasionally beating SenseBERT, specifically trained to be better at such tasks.
% \begin{itemize}[leftmargin=*]
%     \item Tokenization matters - Models employing \textbf{SentencePiece + Unigram subword tokenization tend to demonstrate overall better performance and sense discrimination}.
%     \item Despite their overall poor performance (c.f. Appendix \ref{app:Autoregressive Models}), \textbf{CLMs exhibit a strong degree of sense awareness} in OOD datasets.
%     \item \textbf{BERT should be used as a starting point for linguistically challenging datasets}, as it shows strong sense discrimination in closed domains occasionally beating SenseBERT, specifically trained to be better at such tasks.  
% \end{itemize}

\subsubsection{Architecture Examination}

\paragraph*{Motivation} As discussed previously (sec. \ref{sec:Introduction}), scaling LLMs does not always lead to SOTA performance on every task. Accordingly, we test whether \textcolor{blue}{\textit{scaling LMs is a definitive solution for zero-shot generalization or not}}. We also look at how pre-processing decisions impact generalization.
%Therefore, we decided to test if scaling laws apply to domain transfer, i.e., whether it was similarly inadequate or not. 

\paragraph*{Experiment Setting} Here, we study how variations in the model architecture impact EQA performance. This includes analysis across varying attention heads, hidden dimensions and number of layers. Also, we look at how tokenization decisions including word masking and text normalization contribute to the performance drop. %Finally, we probe the zero-shot capabilities of a variety of decoder-based models, each built using novel attention mechanisms, cleaner/high-quality corpora and scaled to levels not previously seen before especially for the encoder-class of models.

%\label{sec:Bi-Directional Models}
%\suhang{why do we put another paragraph title here? remove it? - addressed}
As it is computationally intractable to analyze every LLM, we perform a deep dive on one particular model BERT to glean an understanding of the performance drop. Additionally, we use BERT due to the availability of a variety of pre-trained configurations as released by \citet{turc2019well}.

%Ideally, we would want to analyze a range of different architectures to yield stronger conclusions. However, as this is , %given the number of extant language models, 

%To this end, we use , a deep learning architecture which computes token representations by reading the input text from left-to-right and right-to-left (hence bi-directional). 

%Importantly, we wanted a model for which there exists several pre-trained variants across a combination of architecture configurations (see above). Otherwise, it would have become extremely expensive to have to train each model from scratch for our experiments. For this reason 
%These models were developed as a part of a knowledge distillation and quantization study in which it was shown empirically that a simple baseline of directly pre-training smaller models for language modelling is more effective than distilling a randomly initialized student from a larger teacher.

%The idea for our experiment is shown in Figure \ref{fig:bert-models}. 
For each pre-trained variant\footnote{See Appendix \ref{app:BERT_configurations_tested} for each configuration tested.} with a given combination of layers ($L$), attention heads/hidden layer size ($A|H$), we first fine-tune them on SQuAD (Fig. \ref{fig:problem_statement}) and then zero-shot test on the OOD datasets to gauge the impact of model size on performance. %From Fig. \ref{fig:bert-models} we obtain a total of 25 variations. since $A|H=16|1024$ is only available for the $L=24$ model (BERT\textsubscript{LARGE}).

We also investigate how pre-processing decisions impact performance. Under this bracket, we test whether \textit{whole-word-masking} (WWM), i.e., masking ALL of the tokens associated with a word %e.g. BERT tokenizes \textit{Croissant} as [\textit{`cr', `\#\#ois', `\#\#sant'}] and the choice is made to mask EACH token, 
makes any difference over \textit{word-piece-masking} (WPM), i.e., masking some or none of the tokens of a word randomly. It can be reasonably expected that closed-domain datasets will contain information on entities typically not observed in the general domain. %such as \textit{DC-SIGNR} (COVID-QA). 
Thus, we also test whether \textit{case preserving} models perform better than \textit{case-independent} ones since it has been established that models exploit casing to identify entities \cite{das2022resilience}.

%The overall breakdown of models by pre-processing decisions is explained in Fig. \ref{fig:bert-classification}. From this classification, we obtain a total of 29 variations to test - for WPM, there exists 25 uncased models as described before and two cased (only BERT\textsubscript{BASE} and BERT\textsubscript{LARGE}) models; for WWM, we have only two models, both BERT\textsubscript{LARGE} for cased and uncased normalization.

\begin{table}[ht]
\centering
\resizebox{\columnwidth}{!}{%
\begin{tabular}{@{}cccccc@{}}
\toprule
                           & \multicolumn{5}{c}{Dataset (EM | F1)}    \\ \midrule
\multicolumn{1}{c|}{Model} & SQuAD & COVID-QA & DuoRC & TechQA & CUAD \\ \midrule
\multicolumn{1}{c|}{BERT\textsubscript{Base}[L=12 | A=12 | H=768]}   & 81.30 | 88.58 & 20.8 | 39.3   & 54.16 | 66.82 & 2.93 | 6.58 & 2.44 | 5.26 \\
\multicolumn{1}{c|}{BERT\textsubscript{Large}[L=24 | A=16 | H=1024]} & 84.03 | 91.1  & 22.29 | 40.26 & 53.99 | 67.52 & 2.58 | 7.08 & 2.34 | 4.84 \\ \bottomrule
\end{tabular}%
}
\caption{Impact of using Cased Models with WPM.}
\label{tab:cased-wpm}
\vskip -1em
\end{table}

\begin{table}[ht]
\centering
\resizebox{\columnwidth}{!}{%
\begin{tabular}{@{}c|ccccc@{}}
\toprule
Model                     & \multicolumn{5}{c}{Dataset (EM|F1)}                                     \\ \midrule
BERT\textsubscript{Large} & SQuAD       & COVID-QA      & DuoRC         & TechQA      & CUAD        \\ \midrule
Uncased                   & 86.7 | 92.8 & 22.39 | 38.61 & 58.2 | 71.47  & 2.59 | 8.97 & 2.2 | 4.1   \\
Cased                     & 86.7 | 92.9 & 21.25 | 37.74 & 57.29 | 70.49 & 3.23 | 7.6  & 0.43 | 1.66 \\ \bottomrule
\end{tabular}%
}
\caption{Impact of text normalization with WWM.}
\label{tab:WWM}
\vskip -1em
\end{table}

\paragraph*{Findings A: (Word-Piece Masking)} (WPM) Table \ref{tab:closed-domain-scaling-wpm} provide the results of our zero-shot EQA trials for BERT models trained with word-piece masking and using uncased text. As can be seen, BERT follows a consistent trend for ID SQuAD with performance improving across all axes of model size, i.e., $A$, $H$ and $L$. Additionally, it benefits from training on cased-text as both BERT\textsubscript{BASE} and BERT\textsubscript{LARGE} report the best scores on SQuAD in Table \ref{tab:cased-wpm}. 

Curiously, we find that results from DuoRC follow a very similar trend with SQuAD. %While the absolute scores are nowhere near SQuAD, 
We explain this by the fact that as DuoRC samples are drawn from Wikipedia, they align more with BERT's training data which in turn allows performance to scale across each axis of model size. Unfortunately, a similar trend is not observed for the other closed-domain datasets, i.e., scaling $A$, $H$ and $L$ does not always lead to improvements when the datasets differ widely as highlighted in Table \ref{tab:closed-domain-scaling-wpm}. We reason that this behaviour is caused by ID fine-tuning strongly aligning the base model with the domain to the extent that any architectural modifications do not yield appreciable gains on OOD datasets.

\begin{table}[ht]
\centering
\small
\resizebox{0.8\columnwidth}{!}{%
\begin{tabular}{c|c|c|c}
\toprule
L & A | H & \begin{tabular}[c]{@{}c@{}}EM | F1\\ (SQuAD)\end{tabular} & \begin{tabular}[c]{@{}c@{}}EM | F1\\ (OOD)\end{tabular} \\ \hline
8                    &                            & {\color[HTML]{CD9934} 80.68 | 88.38} & {\color[HTML]{34FF34} 21.3 | 38.02}  \\
10                   & \multirow{-2}{*}{12 | 768} & {\color[HTML]{663234} 81.33 | 88.66} & {\color[HTML]{FE0000} 19.47 | 35.45} \\ \hline
                     & 8 | 512                    & {\color[HTML]{CD9934} 79.74 | 87.41} & {\color[HTML]{34FF34} 1.61 | 6.86}   \\
\multirow{-2}{*}{12} & 12 | 768                   & {\color[HTML]{663234} 80.9 | 88.2}   & {\color[HTML]{FE0000} 1.61 | 6.36}   \\ \hline
12                   & 12 | 768                   & {\color[HTML]{CD9934} 80.9 | 88.2}   & {\color[HTML]{34FF34} 2.46 | 4.63}   \\
24                   & 16 | 1024                  & {\color[HTML]{663234} 83.49 | 90.6}  & {\color[HTML]{FE0000} 0.78 | 3.56}  \\ \bottomrule
\end{tabular}%
}
\caption{Impact of scaling $L$ for COVID-QA (top), $A|H$ for TechQA (middle), both for CUAD (bottom); SQuAD scores are in the third  column. See Figure \ref{fig:WPM-uncased} for scores from all configurations.}
\label{tab:closed-domain-scaling-wpm}
\vskip -1em
\end{table}
    
\paragraph*{Findings B: Whole Word Masking} (WWM) There are only two models to examine under this masking strategy, i.e., BERT\textsubscript{LARGE} trained with and without cased text. Comparing scores from Table \ref{tab:cased-wpm} and \ref{tab:WWM} we see that cased-text in combination with whole-word masking leads to improved scores for SQuAD, DuoRC and TechQA. This makes sense as closed domains discuss various entities and have a processing scheme recognizing that is beneficial.

Overall, we find that WWM tends to outperform WPM. Such an observation was also made by \citet{joshi2020spanbert} who found that span (in our case whole words) prediction as opposed to individual tokens is a more challenging task and leads to stronger models. Finally, we see that the uncased variants of this scheme display the best performance overall. We reason that this is because the models are more sensitive to the choice of masking than text normalization, e.g., irrespective of capitalization, terms such as \textit{new york} will convey the same information. 

\paragraph*{Key Takeaways} The key insights here are: (i) Although Bi-directional models as BERT are more suitable for EQA, %\textbf{scaling laws do not apply to it for zero-shot EQA} i.e. 
\textit{it is not guaranteed to see improvements in closed-domains simply by increasing model scale}; (ii) \textit{WWM models should be preferred over WPM models for cross-domain EQA}; (iii) If a WWM variant is unavailable, consider using \textit{uncased models as they tend to display better performance across domains}; (iv) \textit{When dealing with long-context datasets, consider using Bi-directional models over CLMs (c.f. Appendix \ref{app:Autoregressive Models})} as they do not face similar issues as the latter. 
% \begin{enumerate}[leftmargin=*]
%     \item Although Bi-directional models as BERT are more suitable for EQA, %\textbf{scaling laws do not apply to it for zero-shot EQA} i.e. 
%     it is not guaranteed to see improvements in closed-domains simply by increasing model scale.

%     \item \textbf{WWM models should be preferred over WPM models for cross-domain EQA}.

%     \item If a WWM variant is unavailable, consider using \textbf{uncased models as they tend to display better performance across domains}.

%     \item \textbf{When dealing with long-context datasets, consider using Bi-directional models over autoregressive LLMs (c.f. Appendix \ref{app:Autoregressive Models})} as they do not face similar issues as the latter. 
% \end{enumerate}

\subsection{Dataset Perspective}
\label{sec:dataset_perspective}

While architecture and training decisions impact cross-domain performance, they cannot be solely accountable for the ID-OOD performance discrepancy. As evidenced by the closed-domain datasets used in this study (\S Appendix \ref{app:Dataset_Models}), the number of samples, along with their average answer/context length, provide initial clues for the disparity as ID models are unaccustomed to such instances. Therefore, we compare OOD datasets with their ID counterpart to see exactly how different they are. We do this through established \textit{quantitative} measures that capture insights from the entire dataset by giving global feedback rather than a per-sample basis qualitative examination.

\subsubsection{Impact of Dataset Similarity on Transferability}

%\suhang{we should not use some algorithms we use as the subsubsection title. Instead, use what you analyze as the title, e.g., Impact of Dataset Simialrity on Transferability - addressed}

Here, we quantify the disparity between ID SQuAD and OOD datasets using two techniques viz., Force-Directed Algorithm (FDA) \cite{fruchterman1991graph} and dataset embeddings \cite{vu-etal-2020-exploring}. Through these measures, we gauge how different OOD datasets are w.r.t SQuAD which aids us in understanding the performance drop better.

\begin{figure}
    \centering
    \includegraphics[scale=0.44]{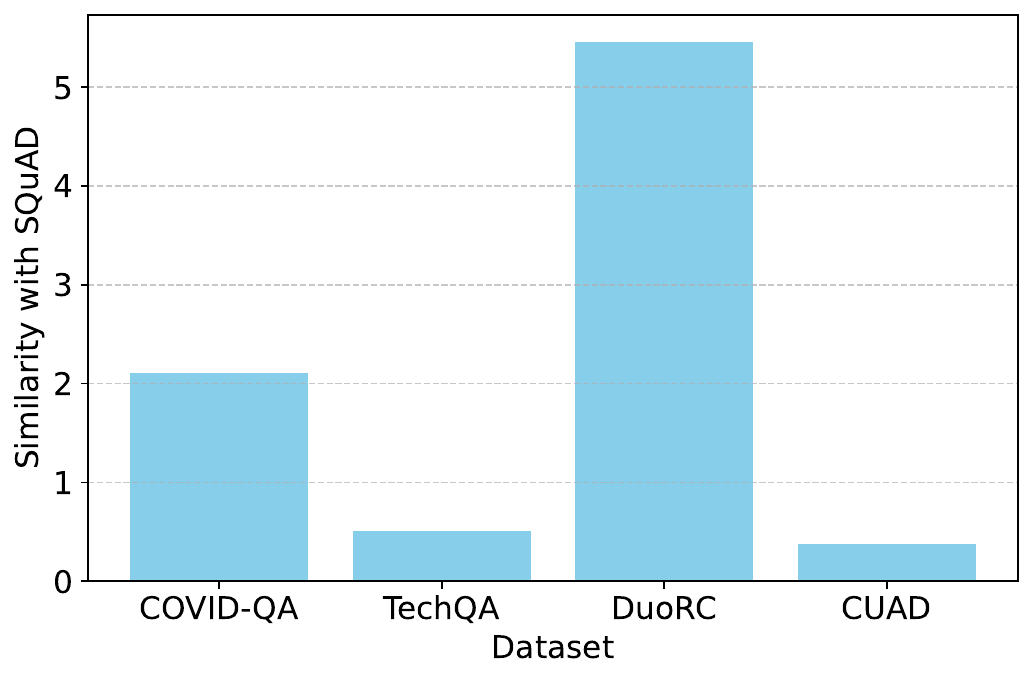}
    \vskip -1em
    \caption{FDA Plot. Each bar represents FDA similarity between SQuAD and the corresponding OOD dataset.}
    \label{fig:FDA}
\end{figure}

\paragraph*{FDA} \citet{su-etal-2019-generalizing} and \citet{talmor-berant-2019-multiqa} study Multi-Task QA and use FDA, a graph construction method, to determine dataset relatedness. According to it, the similarity of a dataset with an OOD one is given as \(\frac{2P_{ij}}{P_j}\), where $D_i$ is a dataset, $P_j$ is the F1 when training and evaluating on $D_j$ and $P_{ij}$ is the F1 when training on $D_i$ and testing on $D_j$. We visualize dataset similarity in Figure \ref{fig:FDA}.

%datasets themselves as nodes and the edges as \textit{springs} between them where similar datasets are closely clustered (pulled) and dissimilar datasets are further (pushed) apart. According to the method, if two datasets are ID their similarity is given by,
%Accordingly, nodes in the graph are connected by edges which are likened to \textit{springs} i.e. closely related nodes are pulled together while unrelated nodes are pushed apart. The ultimate goal with this algorithm is to create a visualization technique capable of displaying multiple entities in a computationally efficient and ``aesthetically–pleasing'' manner.

%\begin{equation}
%    \text{\textit{Similarity}\textsubscript{ID}} = \frac{{P_{ij}}}{{P_j}} + \frac{{P_{ji}}}{{P_i}}
%\end{equation}
\begin{comment}
\begin{equation}
    \text{}
    \label{eq:FDA_OOD_Sim}
\end{equation} 
\end{comment}

\paragraph*{Analysis} Figure \ref{fig:FDA} ranks the OOD datasets in order of similarity with SQuAD as,

{\small \textit{DuoRC > COVID-QA > TechQA > CUAD}}

This makes sense seeing as DuoRC is sampled from Wikipedia, i.e., the same as SQuAD and deals with overall ``simpler'' topics (movie plots) as compared to the other datasets. Dataset characteristics such as longer sample lengths and complex subject matter explain the relative ranking of TechQA and CUAD while COVID-QA strikes a middle ground between the two. This ranking is further reinforced by the overall performance (F1) of the models, i.e.,

{\small
\noindent \textit{DuoRC > COVID-QA > TechQA > CUAD} (BERT) \\
\noindent \textit{DuoRC > COVID-QA > CUAD > TechQA} (RoBERTa)
}

%\item Platypus [Table \ref{tab:dec_zs}] (overall best performing non-domain LLM across all datasets): \textit{COVID-QA (34.25) > DuoRC (30.01) > TechQA > CUAD}

Although there are slight deviations in rank, the overall trend places DuoRC and COVID-QA at the higher end of the similarity spectrum to SQuAD and TechQA/CUAD at the lower end, explained by the models' poor performance.

\paragraph*{\texttt{TEXT}/\texttt{TASK} Embedding}
%\suhang{same here, should not be the tool we use, but what we want to evalaute - addressed}
%Selecting which source task to train the base model on during single-task transfer learning is vital as choosing an unrelated task might lead to negative transfer i.e. hurt performance on the target task. Similarly, for multi-task learning, we want to select the best collection of source tasks which yields the strongest model capable of generalizing well to downstream datasets. Realizing this, 
\citet{vu-etal-2020-exploring} propose two embedding methods to capture \textit{task (dataset) semantics}, i.e., \texttt{TEXT} and \texttt{TASK} embedding (c.f Appendix \ref{app:text_task_embedding} for details). %, the latter of which is inspired by \citet{Achille_2019_ICCV}. 
\texttt{TEXT} embedding captures semantics about the entire dataset while \texttt{TASK} embeddings determine the correlation between different tasks. If the domains/tasks of the two datasets are similar, their \texttt{TEXT}/\texttt{TASK} embeddings will be similar.

We investigate how well each embedding captures dataset semantics. For each dataset, we learn \texttt{TASK} and \texttt{TEXT} embeddings and compute cosine similarity using them between SQuAD and its OOD counterparts. The idea here is to quantify how much ID and OOD datasets differ with the hypothesis being that the two will have, on average, low cosine similarity scores (greater dissimilarity). Following \citet{vu-etal-2020-exploring} we use uncased BERT\textsubscript{BASE} to extract \texttt{TEXT} and \texttt{TASK} embeddings.

Following \citet{turc2019well}, we also establish a non-dense embedding baseline by representing each dataset pair as frequency vectors of the top 100 common unigrams and computing Spearman correlation between them.

%first determining the unigrams common to each pair of SQuAD and OOD dataset and select the top 100 occurring terms based on their frequency counts. This allows us to represent each dataset as a vector of term frequencies using which we compute how (dis)similar they are based on their Spearman correlation values.

\begin{table}[ht]
  \centering
  \resizebox{\columnwidth}{!}{
  \begin{tabular}{@{}lllll@{}}
    \toprule
     & \multicolumn{4}{c}{\textbf{Target Dataset}} \\
    \cmidrule(lr){2-5}
    \multicolumn{1}{c}{\textbf{Embedding Type}} & \multicolumn{1}{c}{COVID-QA} & \multicolumn{1}{c}{TechQA} & \multicolumn{1}{c}{DuoRC} & \multicolumn{1}{c}{CUAD} \\
    \midrule
    Common-Term Frequencies (Sparse) & -0.23 & -0.27 & -0.67 & -0.5 \\
    \midrule
    \multicolumn{1}{c}{\texttt{TEXT} Embedding (Dense)} & 0.9 & 0.82 & 0.92 & 0.86 \\
    \multicolumn{1}{c}{\texttt{TASK} Embedding (Dense)} & 0.77 & 0.64 & 0.86 & 0.63 \\
    \bottomrule
  \end{tabular}}
  \caption{Cosine similarity and Spearman Correlation scores. Each entry indicates the corresponding category score between SQuAD and each OOD dataset. Higher scores indicate greater relatedness.}
  \label{tab:Cosine/Spearman}
  \vskip -1em
\end{table}

\begin{table}[ht]
\centering
\small
\resizebox{\columnwidth}{!}{%
\begin{tabular}{cclll}
\toprule
\multicolumn{1}{c|}{} &
  \multicolumn{4}{c}{Target Dataset} \\ \hline
\multicolumn{1}{c|}{Layer} &
  COVID-QA &
  \multicolumn{1}{c}{TechQA} &
  \multicolumn{1}{c}{DuoRC} &
  \multicolumn{1}{c}{CUAD} \\ \hline
\multicolumn{1}{c|}{1} &
  \cellcolor[HTML]{FFFFFF}{\color[HTML]{00FF00} 0.9} &
  \multicolumn{1}{c}{\cellcolor[HTML]{FFFFFF}{\color[HTML]{00FF00} 0.72}} &
  \multicolumn{1}{c}{\cellcolor[HTML]{FFFFFF}{\color[HTML]{00FF00} 0.97}} &
  \multicolumn{1}{c}{\cellcolor[HTML]{FFFFFF}{\color[HTML]{00FF00} 0.74}} \\
\multicolumn{1}{c|}{2} &
  \cellcolor[HTML]{FFFFFF}{\color[HTML]{00FF00} 0.92} &
  \multicolumn{1}{c}{\cellcolor[HTML]{FFFFFF}{\color[HTML]{00FF00} 0.75}} &
  \multicolumn{1}{c}{\cellcolor[HTML]{FFFFFF}{\color[HTML]{00FF00} 0.97}} &
  \multicolumn{1}{c}{\cellcolor[HTML]{FFFFFF}{\color[HTML]{00FF00} 0.73}} \\
\multicolumn{1}{c|}{3} &
  \cellcolor[HTML]{FFFFFF}{\color[HTML]{00FF00} 0.9} &
  \multicolumn{1}{c}{\cellcolor[HTML]{FFFFFF}{\color[HTML]{00FF00} 0.71}} &
  \multicolumn{1}{c}{\cellcolor[HTML]{FFFFFF}{\color[HTML]{00FF00} 0.96}} &
  \multicolumn{1}{c}{\cellcolor[HTML]{FFFFFF}{\color[HTML]{00FF00} 0.78}} \\
\multicolumn{1}{c|}{4} &
  {\color[HTML]{00FF00} 0.92} &
  \multicolumn{1}{c}{{\color[HTML]{00FF00} 0.69}} &
  \multicolumn{1}{c}{{\color[HTML]{00FF00} 0.97}} &
  \multicolumn{1}{c}{{\color[HTML]{00FF00} 0.74}} \\
\multicolumn{1}{c|}{5} &
  {\color[HTML]{00FF00} 0.91} &
  \multicolumn{1}{c}{{\color[HTML]{00FF00} 0.7}} &
  \multicolumn{1}{c}{{\color[HTML]{00FF00} 0.96}} &
  \multicolumn{1}{c}{{\color[HTML]{00FF00} 0.73}} \\
\multicolumn{1}{c|}{6} &
  {\color[HTML]{00FF00} 0.88} &
  \multicolumn{1}{c}{{\color[HTML]{00FF00} 0.74}} &
  \multicolumn{1}{c}{{\color[HTML]{00FF00} 0.95}} &
  \multicolumn{1}{c}{{\color[HTML]{00FF00} 0.76}} \\
\multicolumn{1}{c|}{7} &
  {\color[HTML]{00FF00} 0.88} &
  \multicolumn{1}{c}{{\color[HTML]{00FF00} 0.69}} &
  \multicolumn{1}{c}{{\color[HTML]{00FF00} 0.94}} &
  \multicolumn{1}{c}{{\color[HTML]{00FF00} 0.72}} \\
\multicolumn{1}{c|}{8} &
  {\color[HTML]{00FF00} 0.89} &
  \multicolumn{1}{c}{{\color[HTML]{00FF00} 0.7}} &
  \multicolumn{1}{c}{{\color[HTML]{00FF00} 0.91}} &
  \multicolumn{1}{c}{{\color[HTML]{00FF00} 0.72}} \\
\multicolumn{1}{c|}{9} &
  {\color[HTML]{00FF00} 0.83} &
  \multicolumn{1}{c}{{\color[HTML]{00FF00} 0.64}} &
  \multicolumn{1}{c}{{\color[HTML]{00FF00} 0.92}} &
  \multicolumn{1}{c}{{\color[HTML]{00FF00} 0.68}} \\
\multicolumn{1}{c|}{10} &
  {\color[HTML]{00FF00} 0.85} &
  \multicolumn{1}{c}{{\color[HTML]{00FF00} 0.7}} &
  \multicolumn{1}{c}{{\color[HTML]{00FF00} 0.94}} &
  \multicolumn{1}{c}{{\color[HTML]{00FF00} 0.68}} \\ \hline
\multicolumn{1}{c|}{11} &
  {\color[HTML]{FF0000} 0.37} &
  \multicolumn{1}{c}{{\color[HTML]{00FF00} 0.69}} &
  \multicolumn{1}{c}{{\color[HTML]{00FF00} 0.84}} &
  \multicolumn{1}{c}{{\color[HTML]{FF0000} 0.22}} \\
\multicolumn{1}{c|}{12} &
  {\color[HTML]{FF0000} 8.16E-12} &
  \multicolumn{1}{c}{{\color[HTML]{FF0000} 1.88E-09}} &
  \multicolumn{1}{c}{{\color[HTML]{FF0000} 3.91E-11}} &
  \multicolumn{1}{c}{{\color[HTML]{FF0000} 7.61E-11}} \\ \hline
\multicolumn{1}{c|}{Avg.} &
  0.77 &
  \multicolumn{1}{c}{0.64} &
  \multicolumn{1}{c}{0.86} &
  \multicolumn{1}{c}{0.63} \\ \bottomrule
\end{tabular}%
}
\caption{Layerwise \texttt{TASK} Embedding similarity against SQuAD. We observe that domain divergence takes place mostly in the last 2 layers.}
\label{tab:task_emb_layerwise}
\vskip -2em
\end{table}

\paragraph*{Analysis} Spearman correlation scores from Table \ref{tab:Cosine/Spearman} show a completely different order than FDA as,

{\small
\textit{DuoRC > CUAD > TechQA > COVID-QA}
}

This is due to count-based vectors failing to capture deeper dataset semantics. For example, CUAD, while drastically different from SQuAD in subject matter, use wording typically found in open-domain documents. As such, relying on unigrams alone is bound to pick up on these characteristics reflected in the overall ranking as above. 

\texttt{TEXT} and \texttt{TASK} embeddings reveal a similar pattern as FDA as,

{\small
\noindent \textit{DuoRC > COVID-QA > CUAD > TechQA} (\texttt{TEXT})\\
\textit{DuoRC > COVID-QA > TechQA > CUAD} (\texttt{TASK})
}

Although there exists a slight difference in order, the overall sequence indicates a strong degree of agreement with BERT and RoBERTa's F1 scores and in turn provides further explanation for the performance discrepancy.

%We provide the average similarity score across all layers' embedding w.r.t SQuAD in Table \ref{tab:Cosine/Spearman} and a layer-wise breakdown in Table \ref{tab:task_emb_layerwise}. 
%\suhang{Table xxx shows xxx - addressed}

\texttt{TASK} embeddings produce layer-by-layer representations. This allows us to investigate fine-grained changes during learning. Table \ref{tab:task_emb_layerwise} shows the similarity scores between each layer's \texttt{TASK} embedding for each OOD dataset w.r.t SQuAD. Similarity scores indicate that the models learn EQA well till layer 10 across each domain. However, \textit{in the last layers is where domain divergence manifests}. In other words, we reason that in the last layers, the signal from the domain/dataset becomes so strong as to overpower what the model learned overall about the task which in turn leads to their observed poor performance. %Although they examine only pre-trained models, our observation is corroborated by \citet{ramesh-kashyap-etal-2021-analyzing} who observe how domain divergence increases by layer.

\subsubsection{Model Perplexity v/s Performance}

While typically used for evaluating language models, \textit{perplexity} (PPL) can be extended to evaluate any dataset by converting the samples into a unified representation akin to any unstructured training corpora (\S Appendix \ref{app:Dataset_Perplexity_Background}). %This is the method behind our experiment here. 
We convert each QA dataset into a corpus by combining all the training contexts and questions into a list of unlabeled samples. Using this converted dataset, we compute a model's PPL on it and correlate it with its performance. The logic here is straightforward; \textit{higher perplexity should correspond to lower performance}. As text in closed domains is qualitatively more complex, it is expected that a model will face difficulty in reasoning over them, leading to higher PPL.

We choose BERT and RoBERTa and two autoregressive LLMs (Platypus and Falcon) for this test. To keep the comparison fair between the two classes of models, we use only the answerable questions from DuoRC, TechQA and CUAD. %While this does not allow for a direct comparison of this test with the other two %(as we use a different subset for BERT and RoBERTa in these trials), 
%the overall analysis is still valid. %We calculate PPL for SQuAD and TechQA using their validation split, DuoRC/CUAD using its test set and training data for COVID-QA. 

\begin{figure}
    \centering
    \includegraphics[scale=0.3]{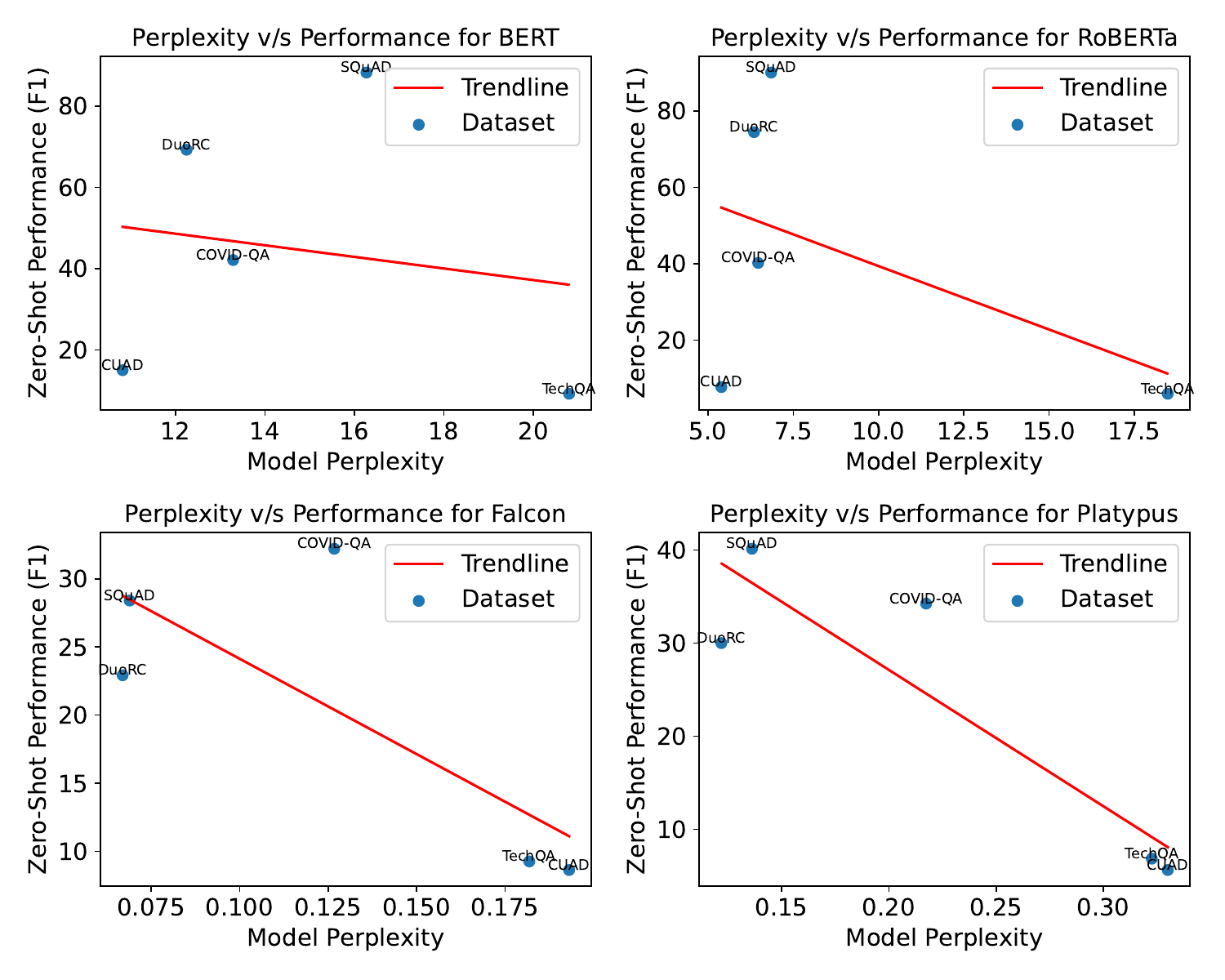}
    \vskip -1em
    \caption{Scatter plot with trend line between model perplexity and performance (F1). Pearson correlation between F1 and PPL. (clockwise from top left) for BERT: -0.17, RoBERTa: -0.48, Falcon: -0.77, Platypus: -0.9.}
    \label{fig:PPLvsPerformance}
\end{figure}

\paragraph*{Analysis} As explained before, our hypothesis is that \textcolor{blue}{\textit{model perplexity on a dataset is inversely proportional to its performance}}. In other words, higher the PPL. lower is its performance. This makes sense since a model's performance is linked with its ability to comprehend the text. Trend lines for all four models (Figure \ref{fig:PPLvsPerformance}) affirm our hypothesis. For example, BERT and RoBERTa report the highest PPL. and corresponding lowest scores for TechQA. A similar observation holds for Falcon and Platypus for CUAD. Although there exist outliers, overall, we see that for datasets with lower PPL. each model shows strong performance. This observation is displayed sharply by the LLMs, as they are overall much better at language modelling, as we can see from the location of SQuAD and DuoRC in the plots for Falcon and Platypus.

%%%%%%%%%%%%%%%%%%%%%%%%%%%%%%%%%%%%%%%%%%%%%%%%%%%%%%%%%%%%%%%%%%%%%%%%

\section{Related Work}
%\suhang{the related work section is too short. You can have a longer version in appendix - addressed}

%Secondly, unlike our setup, none of their datasets are truly ``closed-domains'' i.e. discuss complex subject matter typically not associated with general knowledge. 
%Additionally, they ignore bi-directional models for analysis. 
%gives the model a stronger learning signal than by relying only on their base knowledge (zero-shot). As explained before, fine-tuning
%which is nowhere near the drop experienced by our models %on the datasets we use 
%indicating that 

\textbf{Generalization in LLMs.} LLMs performance on unseen domains \cite{ramponi-plank-2020-neural} remains an active area of study. Recently, \citet{yang-etal-2024-unveiling} and \citet{leng2024towards} examined the effects of fine-tuning on generalization. The issue here is that they examine generalization \textit{after} training, which is not always possible due to data scarcity. \citet{mai2024llms} study LLM generalization on a synthetic domain of ``gibberish'' language. Although novel, their finding's impact on real-world domains remains unclear. 

\noindent \textbf{QA Analysis} Recent studies by \citet{pezeshkpour-hruschka-2024-large} and \citet{khatun2024study} examine the limitations of LLMs for Multiple Choice QA. Although they do not consider EQA, they provide interesting insights such as LLMs being sensitive to the location of answer choices, etc. The work done by \citet{kamalloo-etal-2023-evaluating} is in a similar direction as ours, focusing on the limitations of existing metrics for evaluating extractive or generative QA systems. We find that the closest study to ours is by \citet{miller2020effect} who create new test sets to determine if models trained on SQuAD overfit to it. However, with a reported maximum drop in F1 of 17.4 their datasets are far less challenging to stress test LLMs. Also, they overlook architecture issues to explain the performance gap.

% examined the effects of fine-tuning an LLM on generalization %across a range of tasks and found differences in behavior when trained on generation and classification tasks. 
%while  examine neuron activations in LLMs to determine how they correlate with task performance. %Both studies use at most two LLMs thereby making it difficult to claim the existence of the phenomena across other models. 
%To investigate extreme domain divergence,  examine how well LLMs generalize to a synthetic domain of ``gibberish'' language for learning new ontologies. Although a novel setup, the findings' impact on real world domains remains unclear.

\noindent\textbf{Related Experiments.} \citet{varis-bojar-2021-sequence} examine how Transformers struggle with length generalization if they are trained solely on samples of a given length, potentially indicating overfitting. However, they do not examine other neural architectures as us (c.f. \ref{sec:answer_length}) thus limiting the scope of their claims. %The importance of this test is highlighted by recent works such as \citet{han-etal-2024-lm} who propose methods to adapt LLMs to various context lengths. 
\citet{yenicelik-etal-2020-bert} study how BERT organizes polysemous words in embedding space. However, i) they neglect OOD senses and, ii) while their findings explain how BERT views contexts, they do not provide any actionable insights to using models in such settings as ours.

%investigate a string-edit and translation task and show that 

%The proposed experiment is inspired by two prior studies. First, to determine if Transformers \cite{vaswani2017attention} overfit on training data of certain lengths, \cite{varis-bojar-2021-sequence} investigate two tasks a) a simple string-editing task (whether they are able to learn simple string manipulation operations such as \textit{reverse} and \textit{shift}) and b) natural language translation between English-Czech (their chosen example pair) to see if Transformers trained on samples of a given sequence length can generalize to smaller or longer length sequences. Results on both tests showed that they struggle extremely to generalize to longer or even shorter sequences if they are trained solely on samples of a given length which indicates to a certain degree that these models are overfitting to the training data. Second, \cite{liu2024lost} analyze more recent decoder-only architectures tasked with retrieving information in long-context documents. From their experiments on question-answering and a simple key-value extraction task, they report that the performance shows a sharp decline when the required information is not found it the document extremities i.e. at the start or end. This finding can have deep implications for EQA wherein the answer span can occur anywhere in the context.

%%%%%%%%%%%%%%%%%%%%%%%%%%%%%%%%%%%%%%%%%%%%%%%%%%%%%%%%%%%%%%%%%%%%%%%%

\section{Conclusion}

In this paper, we examine why LMs perform poorly on zero-shot EQA in closed-domains. %The objective was to gain a deeper understanding of how LLMs operate and ultimately determine the cause for their poor generalization. 
We consider reasons from both dataset and model perspective. %, i.e., whether there exists factors in the OOD dataset or models themselves. 
Our findings reveal inadequacies in the current generation of models that need to be addressed to realize true domain generalization. Additionally, we also examine the complexities of OOD datasets which a model needs to be made aware of apriori before they can learn to generalize.

%%%%%%%%%%%%%%%%%%%%%%%%%%%%%%%%%%%%%%%%%%%%%%%%%%%%%%%%%%%%%%%%%%%%%%%%

\section*{Limitations}

For our polysemy tests, a point of concern might be the number of samples in our dataset. While we would have liked to use more instances, we are limited by the size of the domain dataset and consequently the number of samples we can collect for each polysemous term. That said, we believe that our findings are overall still valid as we prioritize sample quality over quantity. Additionally, for the model scale test, while it would be ideal to test even more models, as explained before, it is intractable to test all configurations for every possible model. As such, we decided to thoroughly examine one particular architecture.

\section*{Ethics Statement}

As our study does not deal with sensitive information or involve multiple GPUs for training, we believe that the ethical implications of our work are limited, if any.

\section*{Acknowledgment}
This material is based upon work supported by, or in part by, the Army Research Office (ARO) under grant number W911NF-21-10198 and Cisco Faculty Research Award.

\bibliography{custom}

\appendix

\section{Datasets and Models}
\label{app:Dataset_Models}

Here, we describe all the resources used in this paper, i.e. the models as well as the datasets.

\subsection{Datasets/Domains Studied}

For this study, we utilize five datasets covering a diverse set of domains, as described below. Statistics of the dataset are summarized in Table \ref{tab:dataset-stats}.
\begin{itemize}[leftmargin=*]
    \item SQuAD (Stanford Question Answering Dataset), generally regarded as a benchmark for EQA, is a high-quality open-domain dataset consisting of contexts from Wikipedia and crowdsourced questions-answer pairs based on them.

    \item DuoRC \cite{saha-etal-2018-duorc} is a dataset based on movie plots based on text from Wikipedia and IMDB. In terms of domain closeness, DuoRC is the closest to SQuAD as it includes data from Wikipedia. However, the challenge introduced by DuoRC is its requirement for deeper content understanding since the question and answer are based on different versions of a plot (Wikipedia v/s IMDB) ensuring a lower lexical overlap between the two.
    
    \item CUAD \cite{hendrycks2021cuad} represents the legal domain. Having the longest context length, CUAD is a collection of commercial contracts for legal document understanding.
    
    \item COVID-QA \cite{moller-etal-2020-covid} was developed to enable question answering for COVID-related queries. It is a collection of answerable questions only based on research articles sourced from the CORD-19 dataset \cite{wang-etal-2020-cord}.
    
    \item TechQA \cite{castelli-etal-2020-techqa} is a dataset developed by IBM for question answering in the technical customer support domain. Its subject is not only much different than SQuAD, but the overall language, style and number of samples make this a very challenging dataset. 
\end{itemize}

\begin{table*}[ht]
\centering
\resizebox{\textwidth}{!}{\begin{tabular}{@{}lcccccccccccc@{}}
\toprule
\multirow{8}{*}{\textbf{Dataset}} & \multicolumn{4}{c}{\textbf{Train}} & \multicolumn{4}{c}{\textbf{Validation}} & \multicolumn{4}{c}{\textbf{Test}} \\ \cmidrule(lr){2-5} \cmidrule(lr){6-9} \cmidrule(l){10-13} 
 & \begin{tabular}[c]{@{}c@{}}Average\\ Question\\ Length\end{tabular} & \begin{tabular}[c]{@{}c@{}}Average\\ Context\\ Length\end{tabular} & \begin{tabular}[c]{@{}c@{}}Average\\ Answer\\ Length\end{tabular} & \begin{tabular}[c]{@{}c@{}}Number\\ of\\ Records\end{tabular} & \begin{tabular}[c]{@{}c@{}}Average\\ Question\\ Length\end{tabular} & \begin{tabular}[c]{@{}c@{}}Average\\ Context\\ Length\end{tabular} & \begin{tabular}[c]{@{}c@{}}Average\\ Answer\\ Length\end{tabular} & \begin{tabular}[c]{@{}c@{}}Number\\ of\\ Records\end{tabular} & \begin{tabular}[c]{@{}c@{}}Average\\ Question\\ Length\end{tabular} & \begin{tabular}[c]{@{}c@{}}Average\\ Context\\ Length\end{tabular} & \begin{tabular}[c]{@{}c@{}}Average\\ Answer\\ Length\end{tabular} & \begin{tabular}[c]{@{}c@{}}Number\\ of\\ Records\end{tabular} \\ \midrule
SQuAD & 59.57 & 754.36 & 20.15 & 87,599 & 60.01 & 778.98 & 18.73 & 10,570 & - & - & - & - \\
DuoRC & 40.14 & 3,801.48 & 14.32 & 60,721 (26,633) & 39.97 & 3,837.77 & 14.38 & 12,961 (5,780) & 39.88 & 3,763.95 & 14.27 & 12,559 (3603) \\
COVID-QA & 58.54 & 32,082.28 & 93.42 & 2,019 & - & - & - & - & - & - & - & \\
CUAD & 254.96 & 64,684.51 & 131.48 & 22,450 & - & - & - & - & 260 & 46,848.19 & 120.07 & 4,182 \\
TechQA & 270.88 & 51,452.02 & 269.85 & 600 & 286.77 & 92,629.53 & 156.79 & 310 (9) & - & - & - & - \\ \bottomrule
\end{tabular}}
\caption{Dataset Statistics. Apart from number of records, all entries are text lengths in terms of average number of characters. `-' indicates unavailable dataset split. Numbers in parentheses for TechQA and DuoRC indicate the number of samples for which the answer span is not present \textbf{exactly} in the context for reasons such as inconsistent spaces, capitalization, word inflection (form) variation, etc.}
\label{tab:dataset-stats}
\end{table*}

\subsection{Models Tested}
We test various neural architectures categorized as Non-Transformer and Transformer-based.

\noindent\textbf{Non-Transformer}: We examine two models based on either recurrent or convolution networks. 
\begin{itemize}[topsep=2pt,partopsep=2pt,leftmargin=*]
\setlength{\parskip}{2pt}
\setlength{\itemsep}{0pt}
            \item BiDAF (Bi-Directional Attention Flow) \cite{seo2017bidirectional} is a hierarchical recurrent model (LSTM) that captures fine-grained context and question semantics through word/character embeddings and left-to-right/right-to-left attention.
            \item QANet \cite{DBLP:conf/iclr/YuDLZ00L18} uses unidirectional attention and enables parallelism/scale through several layers of convolution.
\end{itemize}
\textbf{Transformers-Based} \cite{vaswani2017attention}: We look at popular models built using the encoder or decoder part of the transformer.
        \begin{itemize}[topsep=2pt,partopsep=2pt,leftmargin=*]
            \item \textbf{Encoders}: Architectures that fall in this category perform Masked Language Modelling (MLM). This means that during training, the model has access to the entire input sequence, at each step of processing, and is tasked with predicting randomly replaced (\textit{masked}) tokens in the input. In this paper, we explore the following MLM style models,
                \begin{itemize}
                    \item BERT \cite{devlin-etal-2019-bert} is one of the first language models built upon the Transformer-encoder, which showed strong performance across a range of tasks on the GLUE benchmark \cite{wang-etal-2018-glue}. Various \textit{domain-specific} checkpoints of BERT are also tested in this paper to evaluate the impact of further ID pre-training. For the biomedical and technical domain, these include, BioBERT \cite{lee2020biobert} (medical domain) and SciBERT (both domains) \cite{beltagy2019scibert}; for the legal domain, these include, FinBERT \cite{araci2019finbert} and LegalBERT \cite{chalkidis-etal-2020-legal}.
        
                    \item SenseBERT \cite{levine2020sensebert} is pre-trained to predict word senses derived from the English WordNet \cite{fellbaum1998wordnet}. We use this model during our semantic similarity trials (c.f. \ref{sec:Examining Polysemy of Domain Terms}) to determine the impact of this training objective on closed-domain sense discrimination.
                
                    \item RoBERTa \cite{liu2019roberta} optimizes BERT by removing its \textit{Next Sentence Prediction} objective, adding dynamic masking and training over a larger corpus of data for more number of epochs.
                \end{itemize}
    
            \item \textbf{Decoders} (c.f. \ref{tab:dec_zs}): Models that perform causal language modelling (CLM) are termed as decoder-based or autoregressive language models. Such models predict future word(s) based on the preceding context. Here, the attention head allows the model to look only at prior tokens (unidirectional), unlike MLM models. 
                \begin{itemize}
                    \item We use four of the latest CLMs for our experiments, i.e., Falcon \cite{almazrouei2023falcon}, Platypus \cite{lee2023platypus}, Gemma \cite{team2024gemma} and Mistral \cite{jiang2023mistral}. Each model uses various advancements in LLM technology such as Grouped-Query Attention \cite{ainslie-etal-2023-gqa}, Low-Rank Adaptation \cite{hu2022lora}, etc. The most important factor for these models, however, is their training data which undergoes meticulous filtration to ensure high quality, such as the Refined-Web corpus \cite{penedo2023refinedweb} for Falcon. As above, we use various domain-specific checkpoints, as applicable. For the medical domain, this includes, MedAlpaca \cite{han2023medalpaca} and BioMistral \cite{labrak2024biomistral}; AdaptLLM \cite{cheng2024adapting} for the legal domain and Phi-2 \footnote{\url{https://huggingface.co/microsoft/phi-2}} for the technical domain.
                \end{itemize}
    \end{itemize}

\section{Categorical Answer Length Analysis}
\label{app:Categorical_Answer_Length_Analysis}

\begin{figure*}
    \centering
    \includegraphics[scale=0.6]{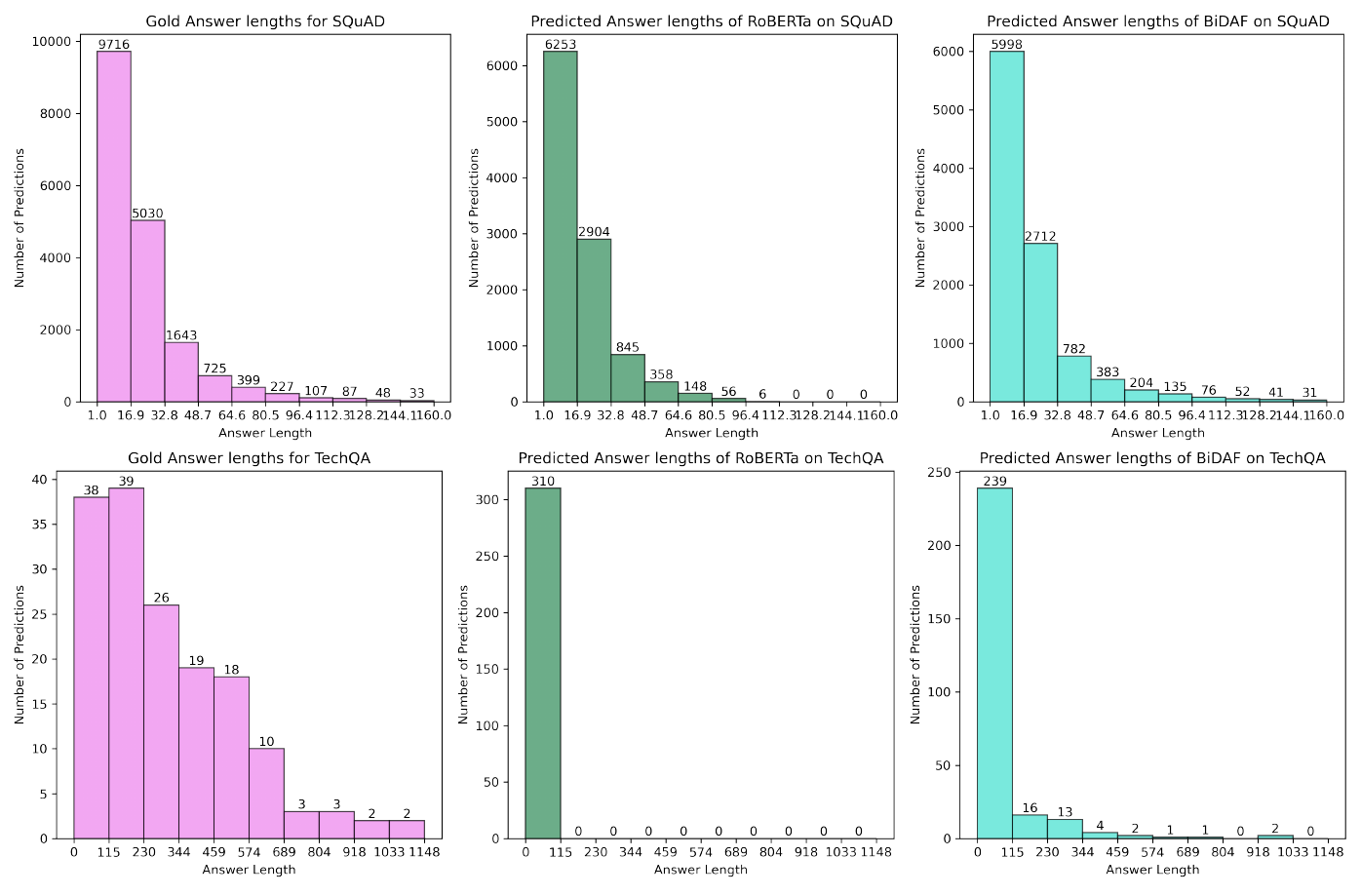}
    \caption{Answer length distribution for BiDAF and RoBERTa on SQuAD (top) and TechQA (bottom).}
    \label{fig:answer-length-distribution}
\end{figure*}

Additionally, we look at how the predicted answer length distributions align with the gold span distribution for two datasets, SQuAD and TechQA for BiDAF and RoBERTa. We chose SQuAD as it is the main dataset on which the models are trained, TechQA as it has the longest average gold span, RoBERTa as it is a better model overall than BERT and BiDAF as observed to be the better of the two non-transformer models. We plot histograms for this test, which are shown in Figure \ref{fig:answer-length-distribution}. The x-axis shows the length ranges of the gold spans for either dataset, and the y-axis shows how many answers fall within each range bucket. We use the same x-axis for all plots for a given dataset to determine the number of predictions falling within the corresponding gold range buckets.

Looking at the fine-grained answer length distribution in Figure \ref{fig:answer-length-distribution} we get a better understanding of why there is poor generalization in MLM style models. For SQuAD, we see that both RoBERTa and BiDAF approximately mimic the gold span answer length distribution. RoBERTa of course performs better than BiDAF owing to its superior architecture and being trained on much more aligned data. However, for TechQA, RoBERTa does not show the same distribution. It is completely left-skewed, with all the predictions falling under 115 characters. On the other hand, BiDAF, despite also being left-skewed, shows a more spread-out distribution. Two of its answers fall in the 918-1033 character range, the same as that in the gold distribution. While RoBERTa cannot break the 30-character length mark for TechQA, BiDAF manages an average of 4k characters \footnote{Longer spans cannot be shown in the plot since they exceed the gold limit (only 278 out of 310 samples are shown in the plot).} (Table \ref{tab:ans-length}). 

\section{Benefits of encoder models (BERT)}

Despite falling out of favour \footnote{\url{https://www.deepset.ai/blog/the-definitive-guide-to-bertmodels}} instead of newer autoregressive models, BERT and its variants have the following advantages,  

\begin{enumerate}
    \item \textit{Shorter training times}, e.x. Pre-training GPT-1 \cite{radford2018improving} took 1 month across 8 GPUs \footnote{\url{https://openai.com/research/language-unsupervised}} v/s BERT\textsubscript{BASE} took 4 days on 16 TPU's \cite{devlin-etal-2019-bert}.

    \item \textit{Smaller model size} e.g. BERT\textsubscript{BASE} has 110M parameters \cite{devlin-etal-2019-bert} v/s GPT-1 has 117M parameters \cite{radford2018improving}. 

    \item \textit{Being more suitable for information extraction tasks} such as Named Entity Recognition \cite{deusser2023informed} (although not always as shown in \cite{sarrouti2022comparing} for biomedical relation extraction, where encoder-decoder models can occasionally top encoder-only models) and span detection (EQA) \cite{xu2021attention, mallick2023adapting}.

    \item \textit{We still find instances of bidirectional language modelling being used in innovative ways}, such as \cite{li2023geolm} who propose an encoder-only model to link text modality with geospatial content. 
\end{enumerate}

\subsection{BERT configurations tested}
\label{app:BERT_configurations_tested}

\begin{figure}
    \centering
    \includegraphics[scale=0.7]{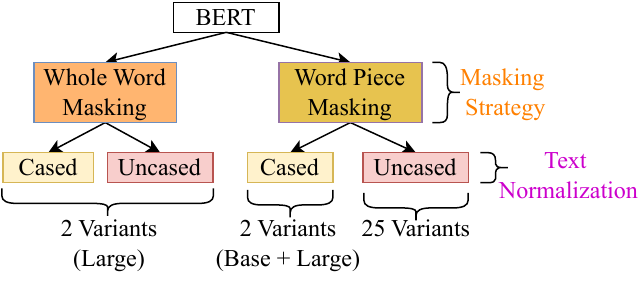}
    \caption{Classification of all BERT models tested for zero-shot EQA based on pre-processing choices. Overall, this gives us 25 (Fig. \ref{fig:bert-models})+2+2=29 variations to test.}
    \label{fig:bert-classification}
\end{figure}

\begin{figure}
    \centering
    \includegraphics[scale=0.7]{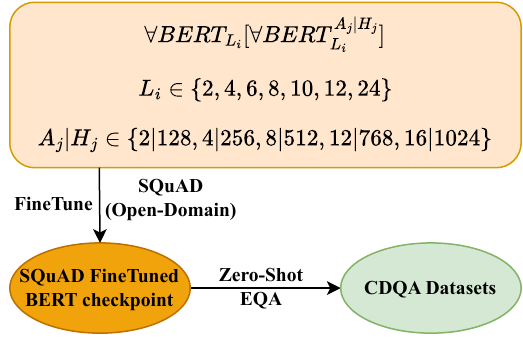}
    \caption{Testing various configurations of BERT to see how they impact zero-shot EQA performance. Note, $A|H = 16|1024$ is only available for $L=24$.}
    \label{fig:bert-models}
\end{figure}

\begin{figure*}
    \centering
    \includegraphics[scale=0.35]{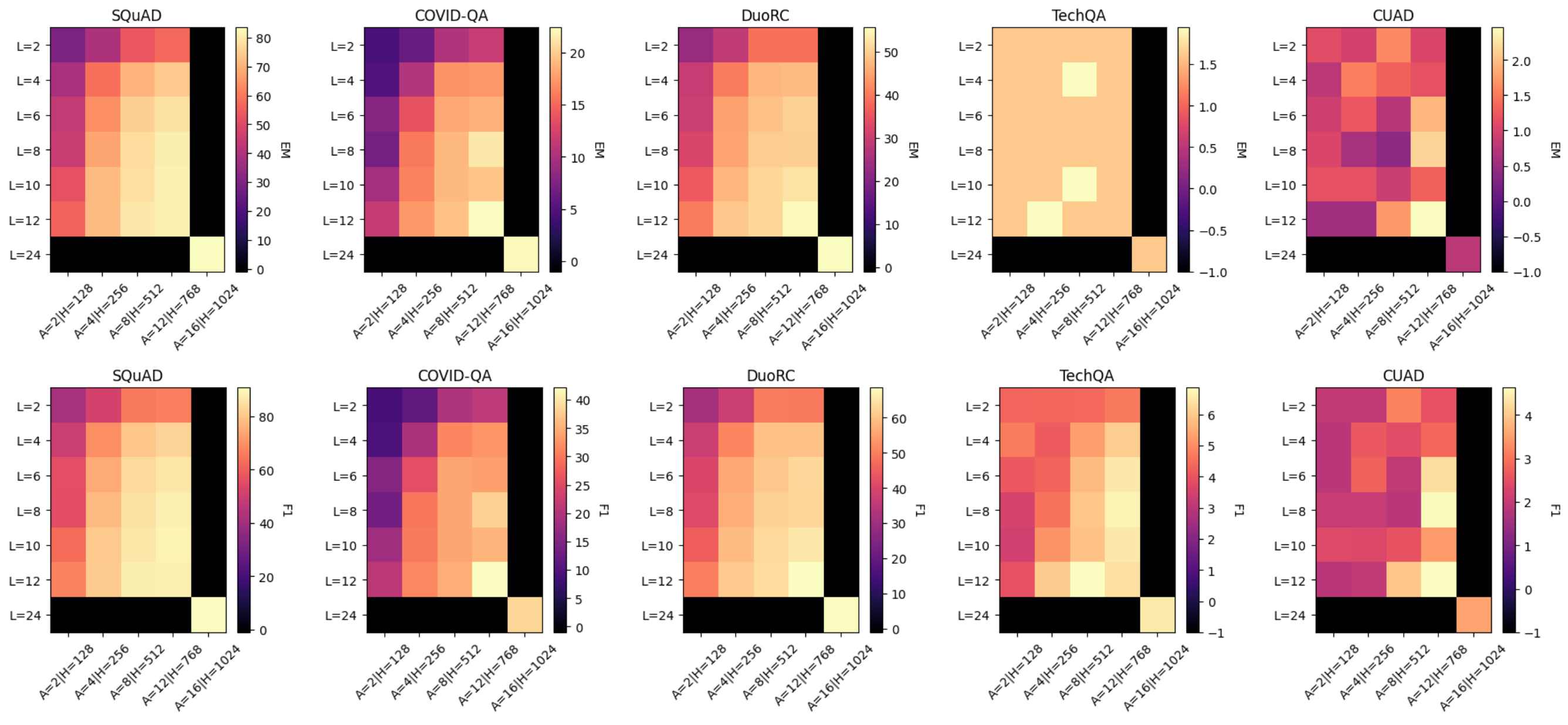}
    \caption{Impact of scaling number of layers (L), attention heads (A) and layer dimension size (H) on EQA generalization for uncased BERT with word-piece masking. EM scores (top) | F1 scores (bottom).}
    \label{fig:WPM-uncased}
\end{figure*}

Figures \ref{fig:bert-classification} and \ref{fig:bert-models} provide an overview of all BERT variations that were tested. Figure \ref{fig:WPM-uncased} provides the range of scores for all configurations of uncased BERT models with word-piece masking. 

\section{Testing ChatGPT}
\label{app:chatgpt}

\begin{figure*}
    \centering
    \includegraphics[width=\textwidth]{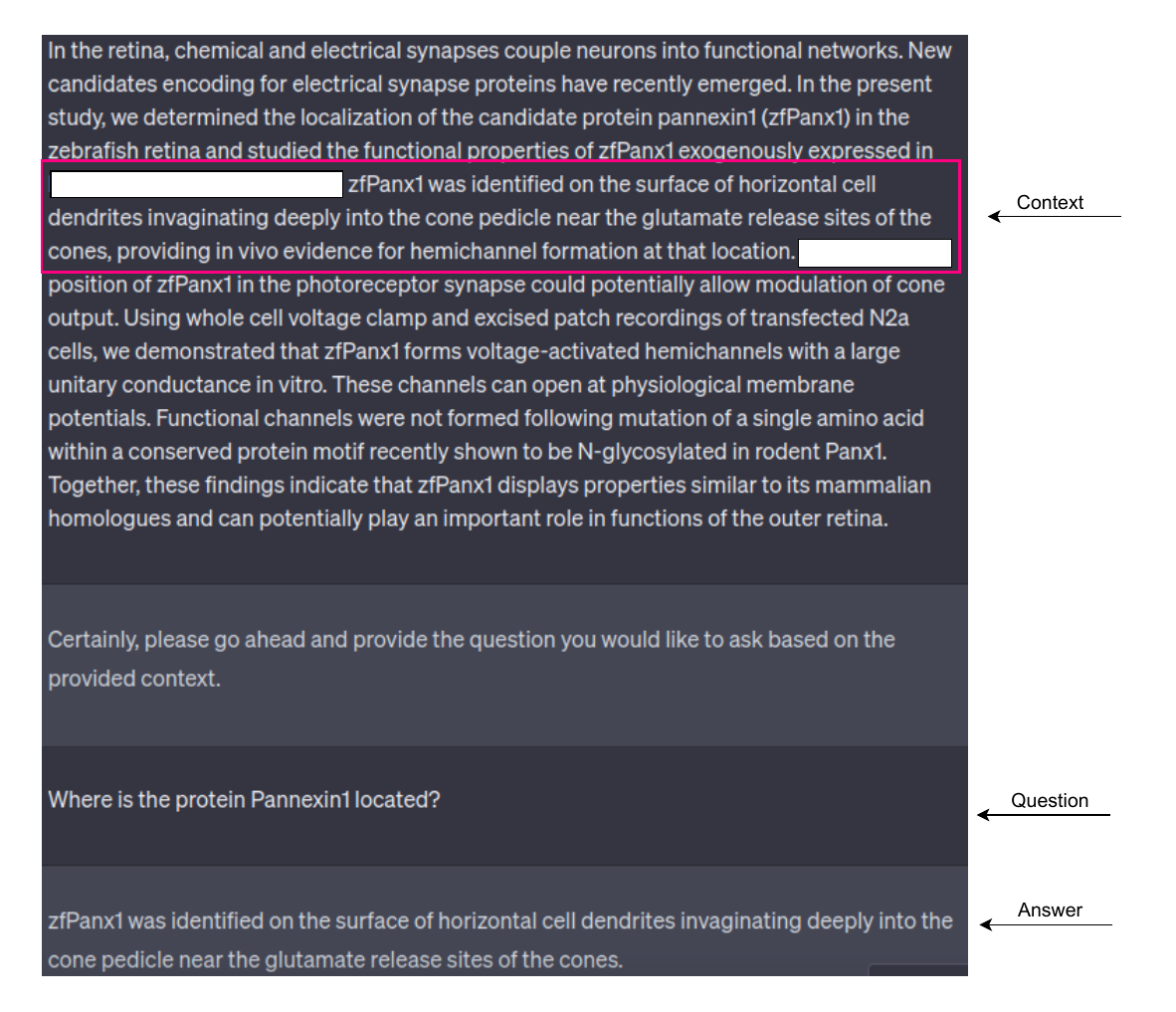}
    \caption{Testing ChatGPT (with GPT-4) on a ``simple'' Biomedical EQA question.}
    \label{fig:chatgpt}
\end{figure*}

We were curious to see how well ChatGPT with either GPT-4 or GPT-3.5 was able to perform zero-shot EQA. We select a random sample from a biomedical dataset, BioASQ \cite{tsatsaronis2015overview} as it has shorter contexts, to test ChatGPT. In Figure \ref{fig:chatgpt}, for the given question, against the true answer of \textit{zfPanx1 was identified on the surface of horizontal cell dendrites invaginating deeply into the cone pedicle near the glutamate release sites of the cones, providing in vivo evidence for hemichannel formation at that location.}, GPT-4 almost identifies the correct span while introducing minimal new text (period instead of comma). GPT-3.5 introduces/patches together even more text with the answer, \textit{The protein Pannexin1 (zfPanx1) is located on the surface of horizontal cell dendrites invaginating deeply into the cone pedicle near the glutamate release sites of the cones in the zebrafish retina.} Changing the prompts did not seem to improve performance. Although this is a single example, it goes to show that even the most capable LLMs struggle with span extraction due to their generative nature and tendency to hallucinate.

\section{Instruction Templates and autoregressive LLM testing setup}
\label{app:gemma}

We use the following prompt template across all models, as described in \citet{han2023medalpaca}. 

\begin{mdframed}[linewidth=0.2pt] 
%\centering
\texttt{Context: \{context text\} \\ Question: \{question text\} \\ Answer: <generated text>}
\end{mdframed}

\begin{figure*}
    \centering
    \includegraphics[scale=0.69]{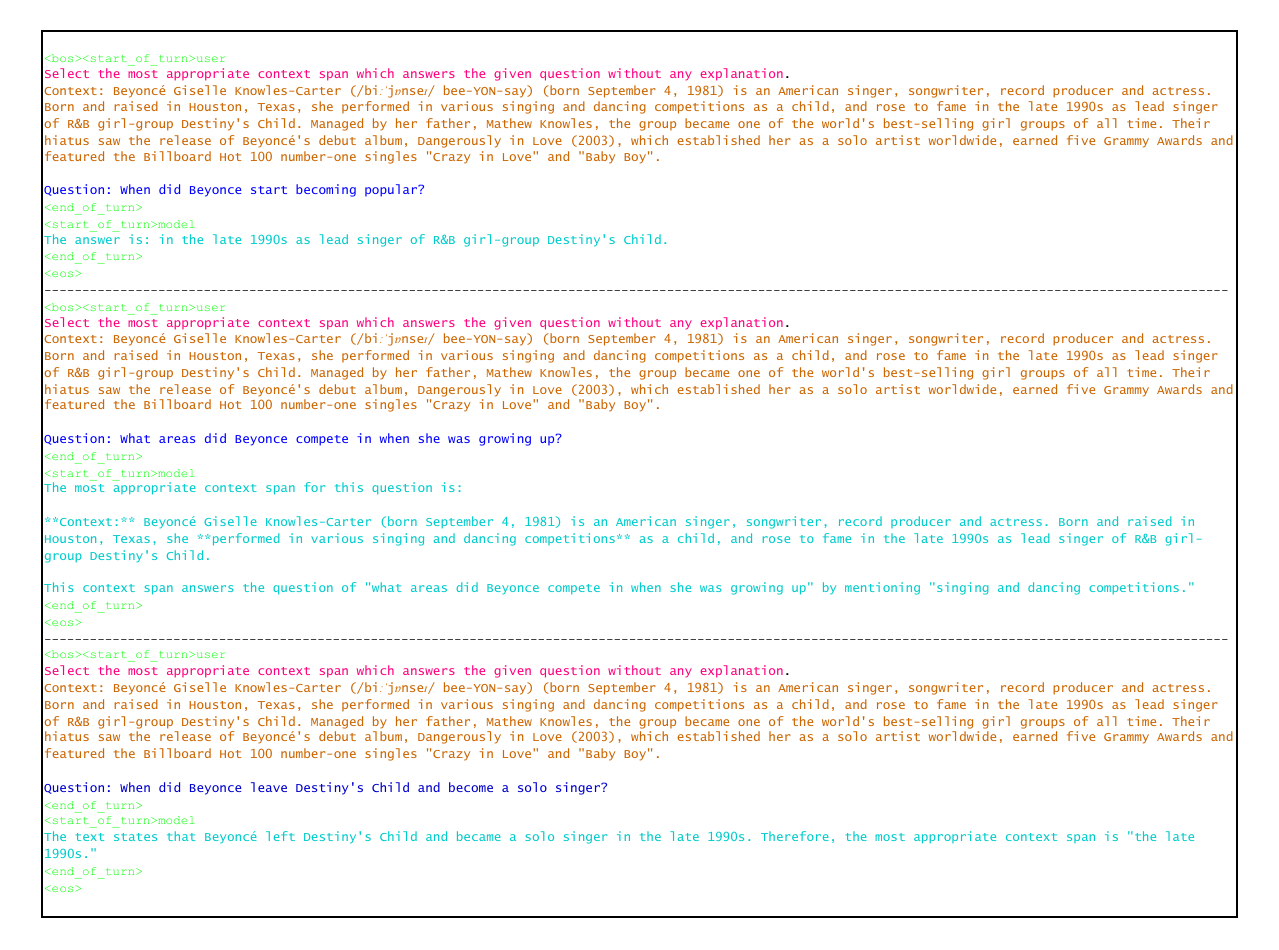}
    \caption{Testing chat template for Gemma. As can be seen, for the recommended template, the model is inconsistent in producing its answer.}
    \label{fig:gemma}
\end{figure*}

The models were prompted in this manner for the following reasons,

\begin{itemize}
    \item Initially, we attempted to format the samples using the instruction tags \footnote{\url{https://huggingface.co/docs/transformers/main/en/chat_templating}} for each corresponding model. However, we found that for certain models such as \texttt{Gemma} \cite{team2024gemma}, even when using the appropriate tags/prompt template, they produce answers in inconsistent formats\footnote{Figure \ref{fig:gemma}}.

    \item It allows us to hold each tested model to a common standard of evaluation, and also provides us with the ease to extract the generated answer automatically by clipping the prompt at a common point (\texttt{Answer:}).
\end{itemize}

In our experiments, the autoregressive models were given the following advantages,

\begin{enumerate}
    \item Retaining answerable questions only - By the very nature of causal training, autoregressive models generate text based on the input prompt. Thus, if the provided context does not contain the answer to the question, they will be penalized for producing an incorrect answer. Thus, we only consider answerable samples for a fair evaluation.
    
    \item Reduced context length - As it has been established that LLMs struggle with processing long contexts \cite{liu2024lost, li2024long}, we truncate the samples from the long context datasets COVID-QA, TechQA and CUAD such that each context chunk is guaranteed to contain the answer while being smaller than the models' maximum input window. SQuAD and DuoRC do not need to be truncated as they have shorter contexts. 
\end{enumerate}

\section{Limitations of autoregressive LLMs for EQA}
\label{app:Autoregressive Models}

The true power of zero-shot learning came into the picture with the release of GPT-3 \cite{brown2020language} a massive 175B autoregressive (or causal) model (trained to predict the next word conditioned only on preceding words) capable of remarkable zero-shot and few-shot learning, i.e., supplied with zero/a few test samples it can perform the target task directly without need of further training. The success of GPT-3 propelled NLP into the LLM era \cite{zhao2023survey} where models are essentially used off-the-shelf for a range of real-world tasks simply by explaining the problem in natural language, a process called \textit{prompting} \cite{liu2023pre}. Harnessing the power of this new feature, we test the generalization capabilities of several state-of-the-art (SOTA) causal models by benchmarking them on our datasets in zero-shot fashion, i.e. directly asking them to answer the question based on the context without further training.

\begin{table}[ht]
\centering
\resizebox{\columnwidth}{!}{%
\begin{tabular}{
>{\columncolor[HTML]{FFFFFF}}c |
>{\columncolor[HTML]{FFFFFF}}c 
>{\columncolor[HTML]{FFFFFF}}c 
>{\columncolor[HTML]{FFFFFF}}c 
>{\columncolor[HTML]{FFFFFF}}c }
Model              & Dataset                                            & EM             & F1             & Predictions in Full Context \\ \hline
\textbf{\textcolor{blue}{MedAlpaca}} & \cellcolor[HTML]{FFFFFF}                           & \textbf{6.79}  & \textbf{39.59} & 384                         \\
Falcon             & \cellcolor[HTML]{FFFFFF}                           & 4.06           & 32.2           & 117                         \\
Platypus           & \cellcolor[HTML]{FFFFFF}                           & 5.25           & 34.25          & 380                         \\
Gemma              & \cellcolor[HTML]{FFFFFF}                           & 2.77           & 18.04          & \textbf{687}                \\
\textcolor{blue}{BioMistral}         & \multirow{-5}{*}{\cellcolor[HTML]{FFFFFF}COVID-QA} & 3.41           & 32.94          & 203                         \\ \hline
Falcon             & \cellcolor[HTML]{FFFFFF}                           & 13.29          & 28.4           & 931                         \\
\textbf{Platypus}  & \cellcolor[HTML]{FFFFFF}                           & \textbf{23.6}  & \textbf{40.14} & 3098                        \\
Gemma              & \cellcolor[HTML]{FFFFFF}                           & 10.78          & 17.45          & \textbf{7364}               \\
Mistral            & \multirow{-4}{*}{\cellcolor[HTML]{FFFFFF}SQuAD}    & 1.2            & 22.25          & 136                         \\ \hline
\textcolor{blue}{AdaptLLM}           & \cellcolor[HTML]{FFFFFF}                           & 0              & 7.38           & 13                          \\
Falcon             & \cellcolor[HTML]{FFFFFF}                           & 0              & 8.63           & 3                           \\
Platypus           & \cellcolor[HTML]{FFFFFF}                           & 0              & 5.64           & 194                         \\
Gemma              & \cellcolor[HTML]{FFFFFF}                           & 0              & 1.45           & \textbf{798}                \\
\textbf{Mistral}             & \multirow{-5}{*}{\cellcolor[HTML]{FFFFFF}CUAD}   & 0 & \textbf{11.6} & 1 \\ \hline
Falcon             & \cellcolor[HTML]{FFFFFF}                           & 0              & \textbf{9.26}  & 0                           \\
Platypus           & \cellcolor[HTML]{FFFFFF}                           & 0              & 6.86           & 0                           \\
\textbf{Gemma}     & \cellcolor[HTML]{FFFFFF}                           & \textbf{0.66}  & 5.11           & \textbf{23}                 \\
Mistral            & \cellcolor[HTML]{FFFFFF}                           & 0              & 8.76           & 0                           \\
{\color[HTML]{000000} \textcolor{blue}{phi-2}} & \multirow{-5}{*}{\cellcolor[HTML]{FFFFFF}TechQA} & 0 & 7.6           & 2 \\ \hline
\textbf{Falcon}    & \cellcolor[HTML]{FFFFFF}                           & \textbf{11.29} & 22.92          & 682                         \\
\textbf{Platypus}  & \cellcolor[HTML]{FFFFFF}                           & 10.1           & \textbf{30.01} & 1707                        \\
Gemma              & \cellcolor[HTML]{FFFFFF}                           & 6              & 10.61          & \textbf{8339}               \\
Mistral            & \multirow{-4}{*}{\cellcolor[HTML]{FFFFFF}DuoRC}    & 4.4            & 29.97          & 517                        
\end{tabular}%
}
\caption{Zero-Shot Decoder Evaluation on all five datasets. Contexts of COVID-QA, TechQA and DuoRC are truncated such that each context chunk always contains the answer. \textcolor{blue}{Blue} indicates ID models while \textbf{bold} is the best performing model.}
\label{tab:dec_zs}
\end{table}

\begin{table}[ht]
\centering
\resizebox{\columnwidth}{!}{%
\begin{tabular}{@{}ccccc|ccc@{}}
\multicolumn{1}{l}{} &
  \multicolumn{1}{l}{} &
  \multicolumn{3}{c|}{Normal Order} &
  \multicolumn{3}{c}{Reverse Order} \\ \midrule
\multicolumn{1}{c|}{Model}     & \multicolumn{1}{c|}{Dataset}                 & CIC  & CIQ  & CIB  & CIC  & CIQ  & CIB  \\ \midrule
\multicolumn{1}{c|}{Falcon}    & \multicolumn{1}{c|}{}                        & 0.26 & 0.31 & 0.2  & 0.12 & 0.18 & 0.09 \\
\multicolumn{1}{c|}{Platypus}  & \multicolumn{1}{c|}{}                        & 0.03 & 0.01 & 0.01 & 0.17 & 0.13 & 0.11 \\
\multicolumn{1}{c|}{Gemma}     & \multicolumn{1}{c|}{}                        & 0.22 & 0.82 & 0.2  & 0.01 & 0.75 & 0.01 \\
\multicolumn{1}{c|}{Mistral}   & \multicolumn{1}{c|}{\multirow{-3}{*}{SQuAD}} & 0    & 0    & 0    & 0    & 0.14 & 0    \\
\multicolumn{1}{c|}{Avg.} &
  \multicolumn{1}{l|}{} &
  \textbf{0.13} &
  \textbf{0.29} &
  \textbf{0.1} &
  \textbf{0.08} &
  \textbf{0.3} &
  \textbf{0.05} \\ \midrule
\multicolumn{1}{c|}{Falcon}    & \multicolumn{1}{c|}{}                        & 0.07 & 0.07 & 0.04 & 0.08 & 0.65 & 0.05 \\
\multicolumn{1}{c|}{Platypus}  & \multicolumn{1}{c|}{}                        & 0.02 & 0    & 0    & 0    & 0.08 & 0    \\
\multicolumn{1}{c|}{Gemma}     & \multicolumn{1}{c|}{}                        & 0    & 0.59 & 0    & 0    & 0.4  & 0    \\
\multicolumn{1}{c|}{Mistral}   & \multicolumn{1}{c|}{\multirow{-3}{*}{DuoRC}} & 0    & 0    & 0    & 0    & 0.1  & 0    \\
\multicolumn{1}{c|}{Avg.}      & \multicolumn{1}{l|}{}                        & 0.02 & 0.17 & 0.01 & 0.02 & 0.31 & 0.01 \\ \midrule
\multicolumn{1}{c|}{MedAlpaca} & \multicolumn{1}{c|}{}                        & 0    & 0.14 & 0    & 0    & 0.15 & 0    \\
\multicolumn{1}{c|}{Falcon}    & \multicolumn{1}{c|}{}                        & 0.05 & 0.28 & 0.02 & 0    & 0.26 & 0    \\
\multicolumn{1}{c|}{Platypus}  & \multicolumn{1}{c|}{}                        & 0    & 0.02 & 0    & 0    & 0.01 & 0    \\
\multicolumn{1}{c|}{Gemma}     & \multicolumn{1}{c|}{}                        & 0    & 0.66 & 0    & 0    & 0.48 & 0    \\
\multicolumn{1}{c|}{BioMistral} &
  \multicolumn{1}{c|}{\multirow{-4}{*}{COVID-QA}} &
  0 &
  0 &
  0 &
  0 &
  0.04 &
  0 \\
\multicolumn{1}{c|}{Avg.}      & \multicolumn{1}{l|}{}                        & 0.01 & 0.22 & 0    & 0    & 0.19 & 0    \\ \bottomrule
\end{tabular}%
}
\caption{Testing instruction following capabilities of each decoder model. Normal Order = Context followed by Question | Reverse Order = Question followed by Context. CIC/CIQ/CIB = Correctly Identified Context/Question/Both. Each score represents the fraction out of 100 randomly selected samples for which the model positively identifies/repeats each component.}
\label{tab:dec_CQID}
\end{table}

In addition to gauging their raw performance \textit{we test two hypotheses} to explain their behaviour, the first of which is linked to their core \textbf{operating objective}. Causal models treat EQA as a standard \textit{language modelling problem} (e.q. \eqref{eq:LLM-objective} \cite{radford2018improving}) where the model (with parameters $\theta$) predicts the future word ($u_i$) given the preceding context ($u_{<i}$) by maximizing the log-probability of the generated sequence.

\begin{equation}
    \sum_{i} \log P(u_{i}|u_{<i}; \theta)
    \label{eq:LLM-objective}
\end{equation}

Contrary to this, bidirectional models, are trained by adding a linear layer on top of the base model and predicting the start and end tokens of the answer span by minimizing the loss \cite{jurafsky2019speech} as shown in e.q. \eqref{eq:bert-eqa-objective} where $P_{start_i}$ is the probability of the $i^{th}$ context token being the start token and similar for the end token \footnote{For details on how the probabilities are calculated, we refer the reader to chapter 14 of \cite{jurafsky2019speech}}.
\begin{equation}
    -\log P_{start_i} -\log P_{end_i}
    \label{eq:bert-eqa-objective}
\end{equation}

Contrasting the language modelling with the start/end token prediction objective, we consider two hypotheses to potentially explain shortcomings in the former for EQA. First, with bidirectional models, we can constrict them to use only context tokens for answer prediction since the linear layer is trained to process only those tokens which means the answer span will \textit{always} come from the context \cite{jurafsky2019speech}. However, with decoder-based models, there is no such constraint, which means that they are free to predict the most likely word(s) from their vocabulary conditioned on the starting text ($Q+C$). Although there have been attempts to remedy this via libraries such as \texttt{lm-format-enforcer}\footnote{\url{https://github.com/noamgat/lm-format-enforcer}} and through additional training on instruction-data \cite{zhou2023controlled}, there is no such requirement \textit{baked} into the makeup of these models. This leads us to test \textit{how many times the generated answer is present in the context}. If the model generates new text rather than using only context tokens, the EM (Exact Match) will decrease considerably, which in turn will lead to its overall poor performance.

The second relates to how the models \textbf{process the input}. Bidirectional models are trained to distinguish between question and context (or any two separate sequences) by using a special $[SEP]$ token. $Q+C$ is then processed as $[CLS] [Q_{i=1}^{n}] [SEP] [C_{j=1}^{m}]$ where $[CLS]$ is a special token used for classification tasks and $Q_i$/$C_j$ are question/context tokens respectively. However, causal models view the entire input as a single sequence without any special separator in between. This leads us to question whether they can \textit{identify the question and context correctly}, which should be considered a basic ability. If they are not able to do so, it could be another explanation for their poor performance. We test this idea by \textit{simply asking the models to repeat the context and questions verbatim} by prompting them as, 

\begin{mdframed}[linewidth=0pt] 
%\centering
\texttt{\textit{Write the context and question exactly}. \\ Context: \{context text\} \\ Question: \{question text\}}
\end{mdframed}

We also test whether the models are sensitive to the location of either component by reversing the order of the context and question. It should be mentioned here that we provide an instruction for this task and use each model's prompt template as they were seen to perform better than in their absence as opposed to the zero-shot setting described before. We explain this behaviour by observing that copying text is qualitatively easier for the model than extraction based on a condition (question). As such, they can follow the instructions much better here. Overall, this will test not only their instruction-following abilities but also reveal a potential flaw in their design, i.e., the inability to identify what portion of the input corresponds to which segment. 

For these experiments, we use only SQuAD, DuoRC and the truncated samples from COVID-QA. This is because repeating the examples from CUAD and TechQA would exceed the model's maximum context window, even if they are truncated. We randomly sample 100 examples from the selected datasets to run our trials and report the fraction of samples that were correctly identified.

We first analyze their cross-domain performance in Table \ref{tab:dec_zs} to investigate the first hypothesis i.e. whether their answers are an exact match with the associated context, since the requirement for EQA is that the answer span must match the context verbatim. Despite having to process less context, poor EM and prediction hit rate (number of times the answer matches the context exactly) indicate that this is a major bottleneck for these models. As discussed previously, this is unsurprising since generation is unconstrained, i.e., conditioned on the seed text, the models are free to predict the next word based on the most probable token in its vocabulary. Overall, we see how serious this issue is since the models reporting the highest hit rates for the more challenging COVID-QA, CUAD and TechQA datasets, could barely break the 40\% mark (percentage of predictions in full context/total number of samples).

Although it is natural to expect that at least the ID models would display the best performance, we find that for COVID-QA, CUAD and TechQA, only MedAlpaca (for COVID-QA) performs the best. For TechQA, this makes sense since phi-2 is a much smaller model (2.7B params) than the others. In the case of CUAD, although AdaptLLM was trained on legal knowledge, it uses LLaMA-1 \cite{touvron2023llama} as the backbone whereas the best-performing model, Mistral, is a much stronger model capable of outperforming the more powerful LLaMA-2 \cite{touvron2023llama2}. 

Results from the context and question identification trials are presented in Table \ref{tab:dec_CQID}. First, model-wise, \texttt{Gemma} displays the most impressive instruction-following abilities as, on average, it reports the most identified samples in either configuration and across domains. This also aligns with the fact that it reported the most number of exact answer predictions for each dataset (c.f. Table \ref{tab:dec_zs}). Second, from the results, it is evident that the location of each component plays an important part, i.e., Normal order or, \texttt{Context} followed by the \texttt{Question}, appears to be the preferred way of formatting samples for EQA. Finally, as expected, each model recognizes the most number of samples for SQuAD displaying again a weakness in generalizing to OOD datasets. Surprisingly, \texttt{MedAlpaca} and \texttt{BioMistral} while being biomedical models are outperformed by \texttt{Gemma}, perhaps owing to its superior instruction tuning. Overall, the takeaways are,

\begin{enumerate}
    \item Although nowhere near good enough for EM, \textbf{LLMs display better performance than Bi-directional models for extremely challenging OOD datasets such as CUAD and TechQA in terms of F1}. Thus, if the dataset can be constrained to only answerable questions, LLMs in zero-shot \textit{could} potentially be a good option.

    \item \textbf{LLMs are sensitive to the location of the context and question in the prompt}. Thus, care should be taken when formatting the samples as it can impact cross-domain performance.
\end{enumerate}

\section{\texttt{TEXT}/\texttt{TASK} Embedding Background}
\label{app:text_task_embedding}

\paragraph*{\texttt{TEXT} Embedding} Each sample is processed by the frozen base model, i.e., without any additional training, and the average of the \textit{pooled} representation from each input sequence stands as the datasets' \texttt{TEXT} embedding. This vector is used to compare datasets across different domains. Here, the specific task is not important, i.e., as long as two datasets belong to the same domain, their \texttt{TEXT} embeddings will be similar.

\paragraph*{\texttt{TASK} Embedding} We provide an intuitive understanding of \texttt{TASK} embeddings and direct interested readers to \citet{Achille_2019_ICCV} and \citet{vu-etal-2020-exploring} for a deeper understanding of related concepts. First, \texttt{TASK} embeddings view the entire model as a real-valued vector with the total number of dimensions equal to the number of model parameters. During training, each dimension of this vector reflects how much a parameter changes or, is affected during backpropagation. In other words, each dimension tracks the gradient of the loss function w.r.t. each parameter. However, for extremely large models, the \texttt{TASK} embedding can become unmanageably high dimensional. Thus, to compress the feature space, they employ the Fisher information matrix \cite{ly2017tutorial} to retain the top-N parameters which have the most impact on model performance and by extension the task itself. For \texttt{TASK} embeddings, the dataset semantics are secondary to the actual task itself, i.e., as long as the two tasks are similar, their corresponding embeddings will be similar even if they deal with different domains. 

\section{Note on Force-Directed Algorithm}

FDA is used to build graphs. However, since we only have a single focal point (SQuAD) and all other datasets are evaluated w.r.t it, it does not make sense to have a graph with just four outgoing edges from a single node. Thus, we use a bar chart for clarity. Also, it should be noted that we kept the values as is without normalization too $[0,1]$, for better visualization. 

\section{Dataset Perplexity Background}
\label{app:Dataset_Perplexity_Background}

When training a language model, PPL is used to gauge how well it understands unseen corpora. Perplexity can be understood from various points. Intuitively, it indicates how \textit{perplexed} or \textit{confused} the model is by the test data. In other words, given a sequence of tokens, PPL measures how likely (probable) the model believes it is grammatically and semantically. Lower PPL on a corpus indicates a well-trained model.

Typically, PPL has been used to describe the performance of causal language models (predicting future tokens given the preceding context). However, the definition can just as easily be extended to MLM (BERT) style models\footnote{The HuggingFace library \cite{wolf-etal-2020-transformers} computes the PPL of both model types in the same way: \url{https://huggingface.co/docs/transformers/en/perplexity}, \url{https://huggingface.co/learn/nlp-course/en/chapter7/3\#fine-tuning-distilbert-with-accelerate}}. Formally, perplexity of a sequence of tokens $X = (x_0 \ldots x_t)$ is given in eq. \eqref{eq:PPL}. Accordingly, it computes the average log probability over the entire sequence of tokens. This value is negated to shift the score to a positive scale and exponentiated for better readability of very small values.
\begin{equation}
    \label{eq:PPL}
    PPL(X) = \exp\left\{-\frac{1}{t} \sum_{i=0}^{t} \log p_{\theta}(x_i | x_{<i})\right\}
\end{equation}

\section{Software}

All of our code and datasets are available at \url{https://github.com/saptarshi059/generalization-hypothesis}.

\section{Why aren't Causal LMs evaluated in section 3.2.1?}

This point was raised by a reviewer during the review process. We provide our clarification as follows, 

\textit{This is a fair question. We can certainly evaluate them. However, the focus was on models that were already trained to predict a certain answer length. Thus, we investigate whether they overfit it or are capable of generalizing to longer spans. As the generation length of autoregressive LLMs can be controlled, we felt their inclusion here was against the point. That said, we do test their other aspects in later sections (App. \ref{app:gemma}/\ref{app:Autoregressive Models}).}

\end{document}